\documentclass{article}
\pdfoutput=1 
\usepackage{arxiv}

\usepackage[utf8]{inputenc} 
\usepackage[T1]{fontenc}    
\usepackage{hyperref}       
\usepackage{url}            
\usepackage{booktabs}       
\usepackage{amsfonts}       
\usepackage{nicefrac}       
\usepackage{microtype}      
\usepackage{lipsum}
\usepackage{graphicx}
\graphicspath{ {./images/} }
\usepackage{amsmath}
\usepackage{algorithm}
\usepackage{algpseudocode}
\usepackage{xr}
\usepackage{float}
\usepackage[round,authoryear]{natbib}

\begin{document}

\title{Meta-learning to Address Data Shift in Time Series Classification}

\author{
Samuel Myren\textsuperscript{ab} \\
    Corresponding Author \\
    \texttt{myrenst@lanl.gov} \\
    P.O. Box 1663 \\
    Los Alamos, NM 87545-0001 \\
    United States\\
   \And
Nidhi Parikh\textsuperscript{a} \\
    P.O. Box 1663 \\
    Los Alamos, NM \\
    87545-0001 \\
    United States\\
   \And
 Natalie Klein\textsuperscript{a} \\
    P.O. Box 1663 \\
    Los Alamos, NM \\
    87545-0001 \\
    United States \\
}

\maketitle

\textsuperscript{a} Los Alamos National Laboratory, Los Alamos, NM, 87545

\textsuperscript{b} Virginia Tech, Blacksburg, VA, 24061

\section*{Abstract}
    Across engineering and scientific domains, traditional deep learning (TDL) models perform well when training and test data share the same distribution. However, the dynamic nature of real-world data, broadly termed \textit{data shift}, renders TDL models prone to rapid performance degradation, requiring costly relabeling and inefficient retraining. Meta-learning, which enables models to adapt quickly to new data with few examples, offers a promising alternative for mitigating these challenges. Here, we systematically compare TDL with fine-tuning and optimization-based meta-learning algorithms to assess their ability to address data shift in time-series classification. We introduce a controlled, task-oriented seismic benchmark (SeisTask) and show that meta-learning typically achieves faster and more stable adaptation with reduced overfitting in data-scarce regimes and smaller model architectures. As data availability and model capacity increase, its advantages diminish, with TDL with fine-tuning performing comparably. Finally, we examine how task diversity influences meta-learning and find that alignment between training and test distributions, rather than diversity alone, drives performance gains. Overall, this work provides a systematic evaluation of when and why meta-learning outperforms TDL under data shift and contributes SeisTask as a benchmark for advancing adaptive learning research in time-series domains.
    
    \textbf{1-6 Keywords: signals, seismology, Reptile, FOMAML, model-agnostic meta-learning, domain generalization}

\section{Introduction} \label{sec:intro}

    Artificial intelligence (AI) has garnered dramatic success across nearly all scientific and engineering domains in recent years. Traditional deep learning (TDL) models (i.e., neural networks trained on big datasets) have driven this progress, performing well when applied to target distributions ($p_{\mathrm{test}}(x,y)$) similar to the training distribution ($p_{\mathrm{train}}(x,y)$). As $p_{\mathrm{test}}(x,y)$ inevitably deviates from $p_{\mathrm{train}}(x,y)$, arising from changes in $p(x)$ and/or $p(y | x)$, TDL model performance often degrades, a phenomenon we call \textit{data shift} \citep{recht_cifar-10_2018, recht_imagenet_2019, miller_accuracy_2021}. 

    While transfer learning offers a path for adapting models to new distributions \citep{zhuang_comprehensive_2021}, it remains challenging due to the opacity of deep architectures, their highly parameterized nature, and the effort required for additional labeling and retraining. In many scientific and engineering domains, such as seismology or mechanical engineering, the problem is exacerbated by the dynamic nature of physical systems, such as sensor drift, environmental variability, and changes in measurement conditions, which induce distributional shifts that degrade model performance over time.
    
    Meta-learning, or “learning to learn,” offers a potential solution. In meta-learning, models are trained across a distribution of related tasks to extract transferable learning strategies that enable rapid adaptation to new tasks using only a few examples. In the context of data shift, meta-learning can treat data from different distributions as distinct but related tasks, allowing for fast adaptation to new, unseen conditions. Where TDL builds specialized models that can risk stagnation, meta-learning builds meta-models capable of generalizing efficiently across datasets and environments.

    In this work, we explore meta-learning’s potential to mitigate data shift in time-series domains from the physical sciences. We introduce a semi-synthetic, task-based seismic time series dataset, SeisTask, designed to emulate realistic variability and data scarcity scenarios. We present an apples-to-apples comparison between meta-learning (namely optimization-based methods of Reptile \citep{chelsea_model-agnostic_2017} and first-order model-agnostic meta-learning (FOMAML; \citealt{nichol_first-order_2018}) and TDL for addressing data shift and uncover that meta-learning’s advantages are significant, but context-dependent. Specifically, meta-learning yields faster learning, improved performance, and more stable fine-tuning when data are scarce or model architectures are relatively small. When data are abundant and architectures are large, TDL and meta-learning perform comparably, and training from scratch may be preferable depending on data characteristics. Finally, we analyze how task diversity influences meta-learning performance and find that emphasizing diversity helps only when it prioritizes tasks similar to the test data.

    Our study identifies the regimes in which meta-learning meaningfully mitigates data shift and provides practical insight for researchers and engineers in signal-based fields seeking to apply adaptive learning strategies. In addition, our contributed SeisTask dataset offers researchers a strong benchmark for the study of adaptive learning strategies in time-series domains, providing a controlled, task-based framework with infused distributional shifts and ground-truth labels for evaluating model generalization and adaptation performance.

    Our research is structured as follows. Section \ref{sec:related_work_and_background} reviews TDL, meta-learning, and related work, and situates our contributions. Section \ref{sec:data} describes the datasets used, including SeisTask and the evaluation-only dataset. Section \ref{sec:data_shift} quantifies model-relevant data shift in SeisTask and defines training and test splits designed to induce distributional shift. Section \ref{sec:results} compares TDL and meta-learning performance across varying architecture sizes and data regimes. Section \ref{sec:diversity} examines the impact of emphasizing task diversity on performance. Finally, Section \ref{sec:discussion} discusses key findings and provides concluding remarks.

\section{Background and Related Work} \label{sec:related_work_and_background}
    
    Traditional deep learning (TDL) aims to learn model parameters $\phi$ that minimize a loss function $\mathcal{L}$ over data $\mathcal{D}=\{(x_i,y_i)\}_{i=1}^n$, where features $x_i \sim p(x)$ and labels $y_i \sim p(y\mid x_i)$, yielding the optimization objective $\arg\min_{\phi}\mathcal{L}(\phi,\mathcal{D})$. The model is traditionally trained under the assumption that future test data are drawn from the same distribution as the training data.
    
    In contrast, meta-learning aims to enable rapid adaptation to new but related data distributions. Meta-learning views the data as a collection of $T$ tasks, $\mathcal{D}=\{\tau_t\}_{t=1}^T$, where each task $\tau_t=\{(x_{it},y_{it})\}_{i=1}^{n_t}$ represents a distinct but related data-generating process. The objective is to learn meta-parameters $\phi$ that capture transferable structure across tasks and can be efficiently adapted to task-specific parameters $\theta_t$ using a small amount of task-specific data. By viewing tasks as distributionally different, the meta-learning perspective posits a compelling approach for enhancing robustness to data shift.
    
    Meta-learning algorithms are typically grouped into three broad categories: black-box-based, optimization-based, and non-parametric approaches \citep{vettoruzzo_advances_2024}. Among these, the optimization-based model-agnostic meta-learning (MAML; \citealt{finn_model-agnostic_2017}) algorithm has become the most widely adopted, demonstrating success in image classification and natural language processing (NLP) \citep{finn_meta-learning_2018, triantafillou_meta-dataset_2020, wang_meta-learning_2020, wu_prototransformer_2021, liu_when_2024}. 
    
    In MAML, task-specific adaptation is performed by updating the meta-parameters $\phi$ using a small support set $\tau_t^{Support}$, for example via $\theta_t = \phi - \alpha \nabla_\phi \mathcal{L}(\phi,\tau_t^{Support})$ and the resulting adapted parameters $\theta_t$ are evaluated on held-out task data to drive meta-parameter updates. However, MAML requires computing second-order gradients during meta-training, which can be computationally expensive and unstable. To address this, first-order variants such as FOMAML and Reptile approximate or avoid second-order updates while retaining competitive performance. In this work, we focus on these first-order methods due to their efficiency, simplicity, and strong generalization behavior. 
    
    Meta-learning has been developed and benchmarked primarily in computer vision and NLP on large, richly labeled datasets \citep{deng_imagenet_2009, lake_human-level_2015, finn_meta-learning_2018, javed_meta-learning_2019, lee_meta-learning_2019, beaulieu_learning_2020, devos_model-agnostic_2021}. Several studies have investigated meta-learning under data shift, but most rely on large datasets and complex model architectures \citep{li_learning_2018, zhang_adaptive_2021, sun_meta-learning_2024, khoee_domain_2024}. While some work has explored the effect of architecture size on meta-learning performance \citep{chen_closer_2019}, little attention has been paid to the interaction between architecture size and training data availability, which is valuable for scientific domains where labeled data are often scarce, noisy, and models are tailored to domain-specific constraints.

    Research applying meta-learning to time-series data remains comparatively limited, with existing work largely focused on classification and forecasting rather than real scientific applications where labeled data are costly and data shift is pervasive \citep{narwariya_meta-learning_2020, wang_meta-learning_2020, talkhi_using_2024, vettoruzzo_advances_2024, wu_meta-learning_2025}. While existing time-series benchmarks can be repurposed for meta-learning \citep{dau_ucr_2018}, most are not explicitly curated in a task-oriented manner with controlled sources of variation and known ground truth, limiting their usefulness for systematic and interpretable evaluation of meta-learning methods. The introduction of SeisTask addresses this gap by providing a domain-relevant, task-based benchmark with realistic variability that is applicable to other time-series domains.

    Because meta-learning operates over a collection of tasks, an important open question is how the diversity of those tasks affects generalization under data shift. Although diversity is often assumed to improve generalization \citep{hospedales_meta-learning_2022}, empirical evidence suggests that similarity between training and test tasks may be more important \citep{setlur_is_2021, kumar_effect_2023, wang_towards_2024}. This issue has not been systematically studied for time-series data in the physical sciences.

    Beyond meta-learning, a range of strategies have been proposed to address data shift, including domain adaptation \citep{singhal_domain_2023}, parameter-efficient adaptation methods such as low-rank adaptation \citep{hu_lora_2022, han_parameter-efficient_2024}, test-time adaptation using unlabeled target data \citep{liang_comprehensive_2025}, and continual learning approaches that mitigate catastrophic forgetting under sequential shifts \citep{wang_comprehensive_2024}. While these methods offer complementary solutions, we focus on meta-learning as a framework for rapid adaptation and evaluate its effectiveness for mitigating data shift in time-series problems from the physical sciences.

\section{Data} \label{sec:data}

    We leverage data scoped from the seismology domain as our case study owing to its strong analogy to other signal-oriented domains in the physical sciences. We focus on the effort of classifying whether or not a seismic waveform contains an event (e.g., an earthquake), which can be challenging in low signal-to-noise (SNR) regimes. Significant research leveraging deep learning for event identification exists \citep{mousavi_deep-learning_2022}, further motivating our study in this domain.
    
    Since our goal is to understand whether meta-learning is useful in dealing with data shift in the context of time series, we create two datasets: 1) SeisTask: an original, task-based, semi-synthetic seismic dataset, and 2) OOD-STEAD: a truly out-of-distribution task-based, real seismic dataset sub-sampled from the publicly available Stanford Earthquake Dataset (STEAD; \citealt{mousavi_stanford_2019}). 
    
    \begin{figure}[h]
        \centering
        \includegraphics[]{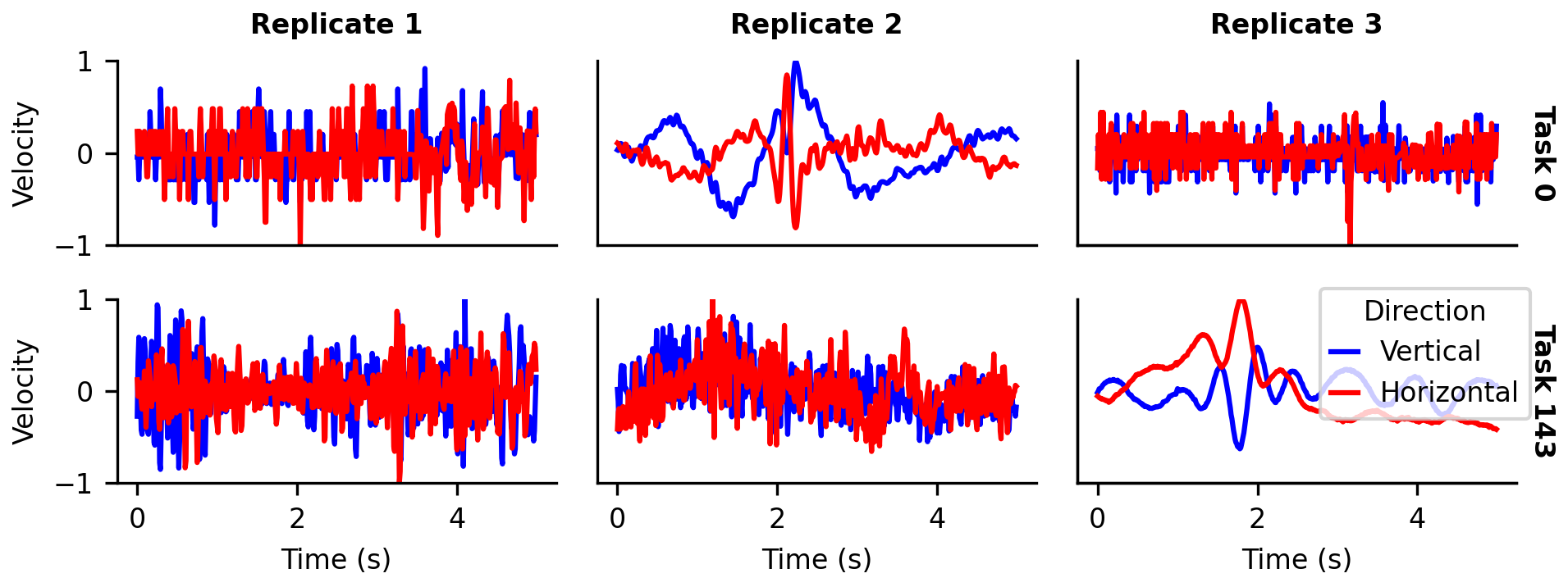}
        \caption{Three replicates of signal waveforms (synthetic signals embedded in real noise) are shown for tasks 0 (top) and 143 (bottom). In the bottom right panel, signal is seen occurring just before 2s, but identifying whether a signal occurs is usually challenging (e.g., in the bottom left panel). Task 0 represents the simulator set at Circles:0, Layers:0, Velocity:Lo, Frequency:Lo, Source:Ricker. Task 143 represents the simulator set at Circles:2, Layers:4, Velocity:Lo, Frequency:Hi, and Source:Gabor. See Appendix \ref{sec:appendix_data} for more details.}
        \label{fig:wf_examples}
    \end{figure}        
    
    \subsection{SeisTask}
        
        SeisTask is an original, semi-synthetic, task-based dataset that enables exploration of data shift and adaptive learning algorithms. It supports a simple, yet challenging, time series classification objective. Tasks in SeisTask are constructed using a design of experiments framework in which earthquake source types and earth substructures are systematically varied using a simulation. As a result, tasks share a common data-generating mechanism, but differ in their signal characteristics. Because SeisTask contains synthetically generated simulation data, it avoids many issues associated with real event data, such as imperfect ground truth and unknown covariate effects.
        
        More specifically, SeisTask is a collection of 243 sub-datasets (or tasks) each containing 420 waveforms. Each task comprises an equal number of waveforms containing a synthetic signal embedded in real noise and waveforms containing only real noise. The noise profiles come from the publicly available Italian dataset for machine learning (INSTANCE; \citealt{michelini_instance_2021}) and share the same distribution. What differentiates the tasks from each other are the characteristics of the synthetic signals, which are generated by a two-dimensional (2D) seismic waveform simulator under certain settings. These settings are governed by a full factorial design of experiments that changes five simulator settings over three levels to make $3^5=243$ distinct groups of signals. As such, the 210 signal waveforms in a given task are obtained through repeated simulator runs at fixed factor levels under random seeds. 

        Each waveform $x_i\in\mathbb{R}^{2\times500}$, representing a two-component (vertical and horizontal directions) 5s long waveform sampled at 100Hz, is paired with a ground truth label $y_i\in\{0,1\}$, denoting whether the waveform contains signal ($y_i=1$) or not ($y_i=0$). We curate the signal waveforms to have an SNR of $1/5$ to enhance the challenge with the classification task. Six examples (or replicates) of signal waveforms from two tasks are displayed in Figure \ref{fig:wf_examples}. Significantly more details regarding the construction of SeisTask and the preparation of the data for modeling are included in Appendix \ref{sec:appendix_data}.

    \subsection{OOD-STEAD: Auxiliary Evaluation Dataset}
        While a subset of SeisTask will be used for training algorithms, we also desire to evaluate our models on true out-of-distribution data. To do so, we create another task set, OOD-STEAD, using real data from STEAD \citep{mousavi_stanford_2019}. Each of the 35 tasks in OOD-STEAD contains 300 signal waveforms and 300 noise waveforms and is differentiated by the SNR range of their signal waveforms. More specifically, from all earthquake waveforms in STEAD that have dataset-provided SNR metadata (not directly comparable to the controlled SNR in SeisTask), we bin the SNR values into 35 bins of increasing SNR. We then randomly select 300 earthquake waveforms from each bin and pair each group of signal waveforms with 300 noise waveforms randomly selected from STEAD without replacement. Finally, we prepare the data for modeling to be consistent with SeisTask (Appendix \ref{sec:appendix_data}).

\section{Quantifying Data Shift in SeisTask} \label{sec:data_shift}

    Before evaluating adaptive learning algorithms, we establish that SeisTask exhibits meaningful data shift from the perspective of the model. For meta-learning to be viable, tasks should be sufficiently shifted from each other to induce model performance degradation, yet similar enough that transferable structure can be learned across tasks. In this section, we quantify these relationships and use them to construct training and test splits that ensure the test tasks are distributionally distinct from those seen during training.

\begin{figure}[H]
        \centering
        \includegraphics[]{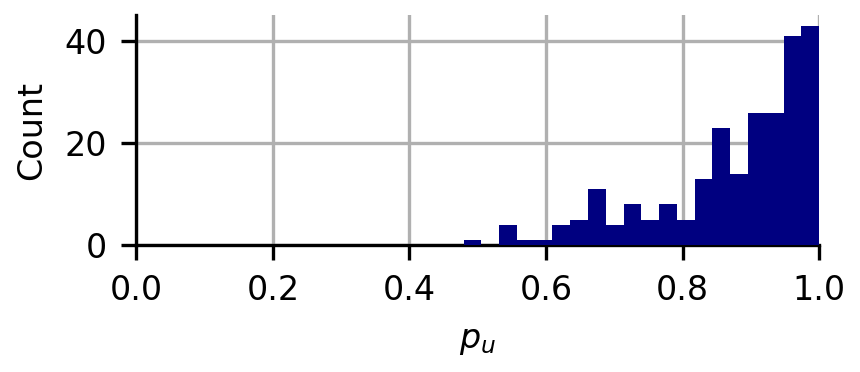}
        \caption{Empirical distribution of the proportion $p_u$ across all tasks in SeisTask. The distribution’s concentration near 1 indicates that models capture task-specific features that improve within-task performance, while its spread reflects transferable information between tasks.}
        \label{fig:p}
    \end{figure}

    To explore task relationships in SeisTask, we first train task-specific models. That is, we train 243 independent models, one for each task, using data exclusively from the task. Then, we evaluate each task-specific model on all other tasks. For each task $u$, we compute the proportion ($p_u$; Eq. \ref{eq:pu}) of the 242 task-specific models \textit{not} trained on task $u$ but evaluated on task $u$, that outperform the model trained and evaluated on task $u$ (Appendix \ref{sec:appendix_datashift}). The histogram showing the empirical distribution of the 243 $p_u$ values is shown in Figure \ref{fig:p}. 
    
    If no model-relevant data shift exists between tasks, then models trained on any task would perform similarly across all others and the distribution of $p_u$ would be uniformly distributed. Conversely, if tasks were distinct with no transferable features, then $p_u$ would approach a point mass at $1$. Because the empirical distribution of $p_u$ is neither uniform nor a point mass, the viability of a meta-learning approach is supported as some features are transferable between tasks while learning task-specific features also remains valuable. 

    While $p_u$ provides insights into the viability of meta-learning for SeisTask, for us to evaluate algorithms in the presence of data shift, we must ensure that the training and test tasks are indeed shifted from the perspective of the model. To quantify data shift between tasks, we compute the \textit{similarity} between any two tasks using linear Center Kernel Alignment \citep{kornblith_similarity_2019}, a metric reflecting the alignment between two task-specific models' latent representations of data (see Appendix \ref{sec:appendix_datashift}).
    
\begin{figure}[H]
        \centering
        \includegraphics[]{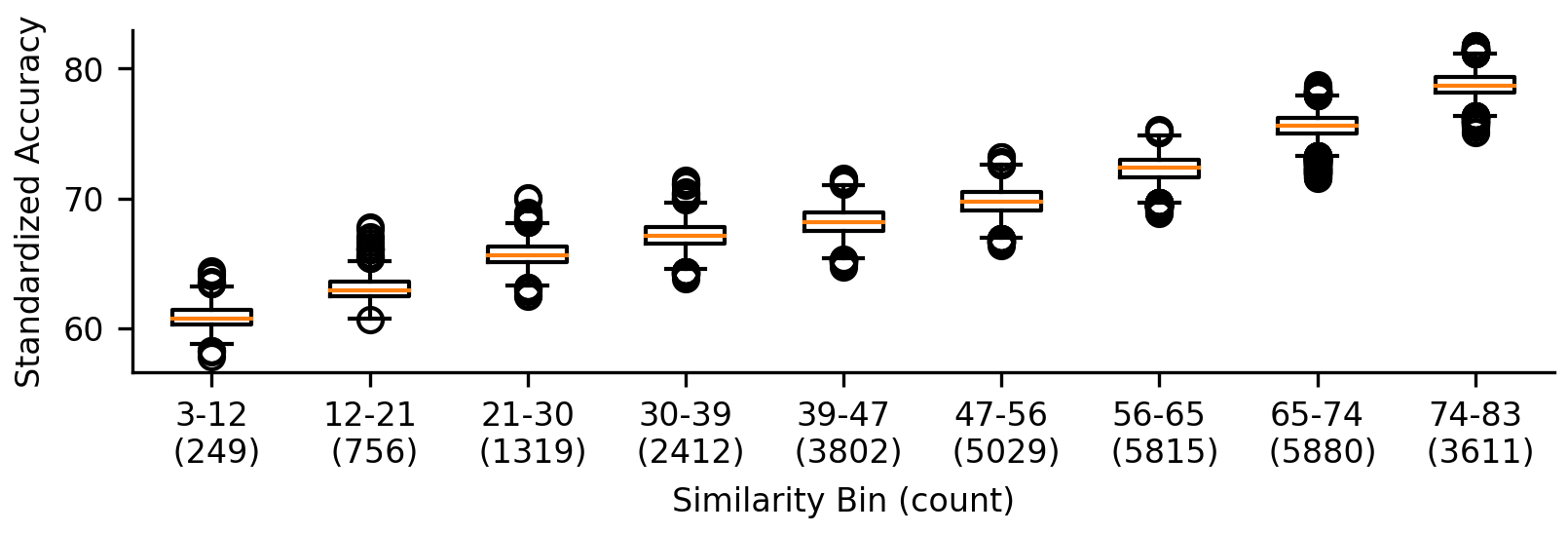}
        \caption{Relationship between cross-task accuracy and task similarity. Each point represents paired tasks binned by similarity. Cross-task accuracy in each bin is standardized to unit variance for display. The positive trend indicates that more similar tasks have better cross-wise performance (when a task-specific model is trained on one task and evaluated on another, and vice-versa), validating similarity as a proxy for relevant data shift.}
        \label{fig:sim_acc}
    \end{figure}

    To verify that similarity captures model-relevant data shift, we inspect the relationship between similarity and cross-task accuracy (the average accuracy of the model trained on task $u$ (or $v$) and evaluated on task $v$ (or $u$)) as shown in Figure \ref{fig:sim_acc}. The linear trend shows that models achieve higher cross-task accuracy when trained and evaluated on more similar tasks. This suggests that similarity provides a reliable, quantitative measure of model-relevant data shift.
    
\begin{figure}[H]
        \centering
        \includegraphics[]{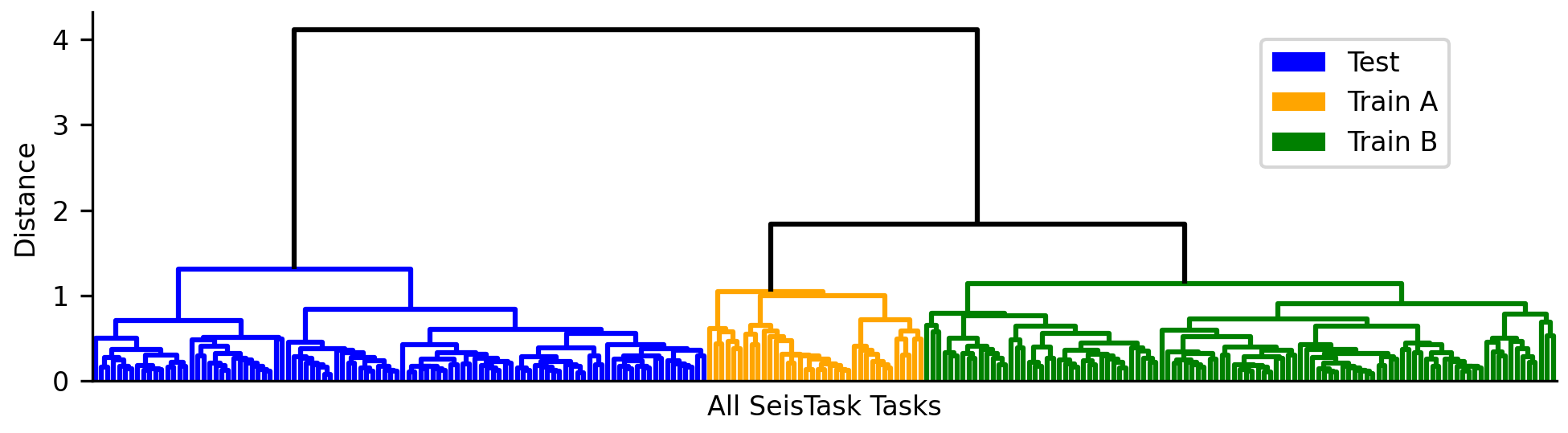}
        \caption{The dendrogram resulting from agglomerative clustering based on task similarity in SeisTask. The tasks on the left side are assigned to the test split, the rest are assigned to the training and validation split. Train A and Train B are distinguished for their relationships to the test split, where Train A is less similar to the test split than Train B.}
        \label{fig:dendrogram}
    \end{figure}

    Using the similarity metric, we construct task splits such that the test split is meaningfully shifted from the training and validation split. To accomplish this, we performed agglomerative clustering with Ward’s linkage criterion \citep{ward_jr_hierarchical_1963} to group related tasks based on similarity. From the resulting dendrogram (Figure \ref{fig:dendrogram}), we assigned the 102 tasks in the initial left branch to the test split and the rest to the training and validation pool. Within this pool, we identified two subgroups, Train A and Train B, based on their relative similarity to the test tasks. As shown in Figures \ref{fig:sim_mat_organized} and \ref{fig:sampling_reduced_heatmap}, Train A is substantially less similar to the test group than Train B, providing insight for post-hoc analysis of how task-sampling strategies influence performance.  

\begin{figure}[H]
        \centering
        \includegraphics[]{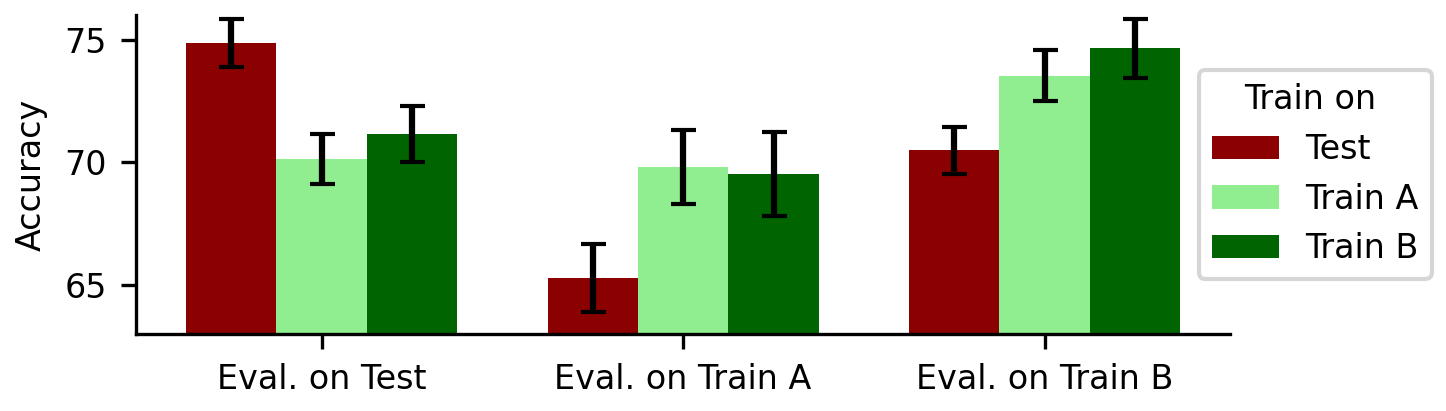}
        \caption{Average accuracy of task-specific models when trained on tasks within a group and evaluated on tasks in other groups. Performance decreases substantially when models trained on Train A or Train B tasks are evaluated on the test tasks, confirming the presence of relevant data shift between the splits.}
        \label{fig:3groups_crosswise}
    \end{figure}

    Finally, Figure \ref{fig:3groups_crosswise} compares performance when task-specific models are trained and evaluated on different task groups defined by the similarity hierarchy. Accuracy drops markedly when models trained on Train A or Train B are evaluated on the test tasks, further supporting that the test tasks are distributionally distinct from the training and validation pool. Having established that SeisTask exhibits relevant data shift, we next evaluate how the algorithms adapt to these shifts during fine-tuning.
        
\section{Evaluating Meta-Learning} \label{sec:results}
    We first outline the experimental framework used to compare meta-learning, TDL, and the divide-and-conquer (D\&C) baseline for addressing data shift. The algorithmic approaches of Reptile and FOMAML leverage the task distinctions during training, whereas TDL ignores them. These three approaches are trained on the designated SeisTask training and validation tasks with 20 model ensembles to quantify variability. The D\&C baseline serves as a benchmark for comparison, where each SeisTask test task (or OOD-STEAD task) is used to train a model from scratch (comprising 10 ensembles) independently of other tasks. We evaluate all models on the SeisTask test tasks and the OOD-STEAD tasks using accuracy as the primary evaluation metric. 
    
    To examine where meta-learning may provide benefits over TDL, we vary the architecture size, amount of training data, and amount of fine-tuning data. We consider four architecture sizes: \textit{mini}, \textit{small}, \textit{big}, and \textit{huge}. We vary the amount of training data $N$, representing half the number of samples per class and per task used to train the base model. Finally, we vary the amount of fine-tuning data $K$, representing the number of samples per class in a test task used to fine-tune the model prior to evaluation on the same task. For fair comparison, we also fine-tune the TDL model to give it the opportunity to adapt like the meta-learning approaches. Fine-tuning was performed for up to 20 gradient steps (or fine-tuning epochs).
    
    To provide an apples-to-apples comparison between algorithms, we use the same architectures, optimizers, learning rate schedules, batch sizes, stopping criteria, and hyperparameters, where possible. More details regarding these nuances are provided in Appendices \ref{sec:appendix_models}, \ref{sec:appendix_algs}, and \ref{sec:appendix_uncertainty}. In the next sections, we provide the results from these experiments and evaluate adaptation consistency and performance in the context of data shift. 

    \subsection{Adaptation Consistency}
\begin{figure}[h]
            \centering
            \includegraphics[]{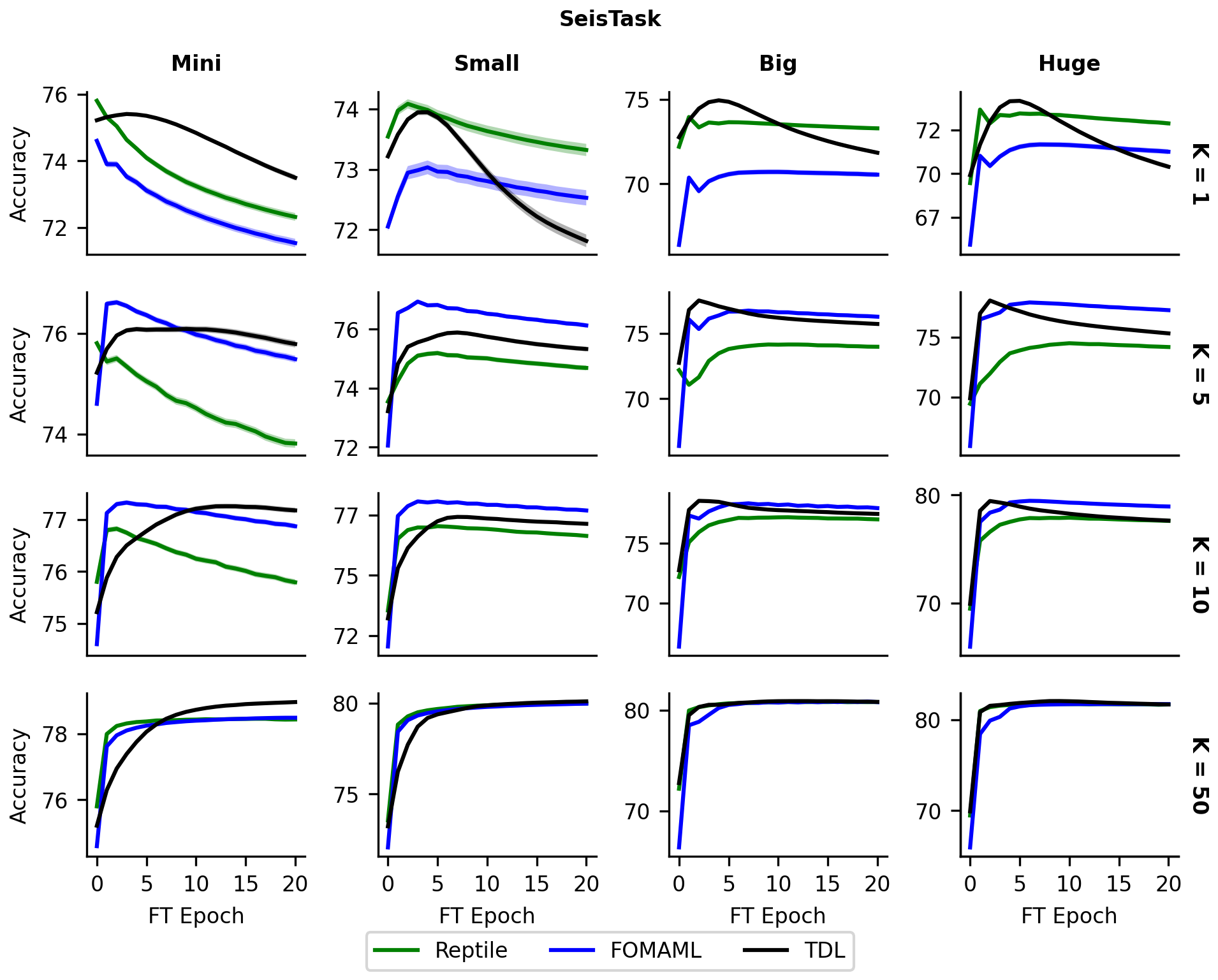}
            \caption{Average performance on \textbf{SeisTask} during fine-tuning, aggregated over training data amount $N$ due to similarity in trends. The x-axis displays the fine-tuning (FT) epoch. Columns are architecture size and rows are the amount of fine-tuning data $K$. Meta-learning algorithms tend to adapt more quickly and guard against overfitting. As $K$ and architecture size grow, algorithmic performance is similar.}
            \label{fig:loss_SeisTask}
        \end{figure}

\begin{figure}[h]
            \centering
            \includegraphics[]{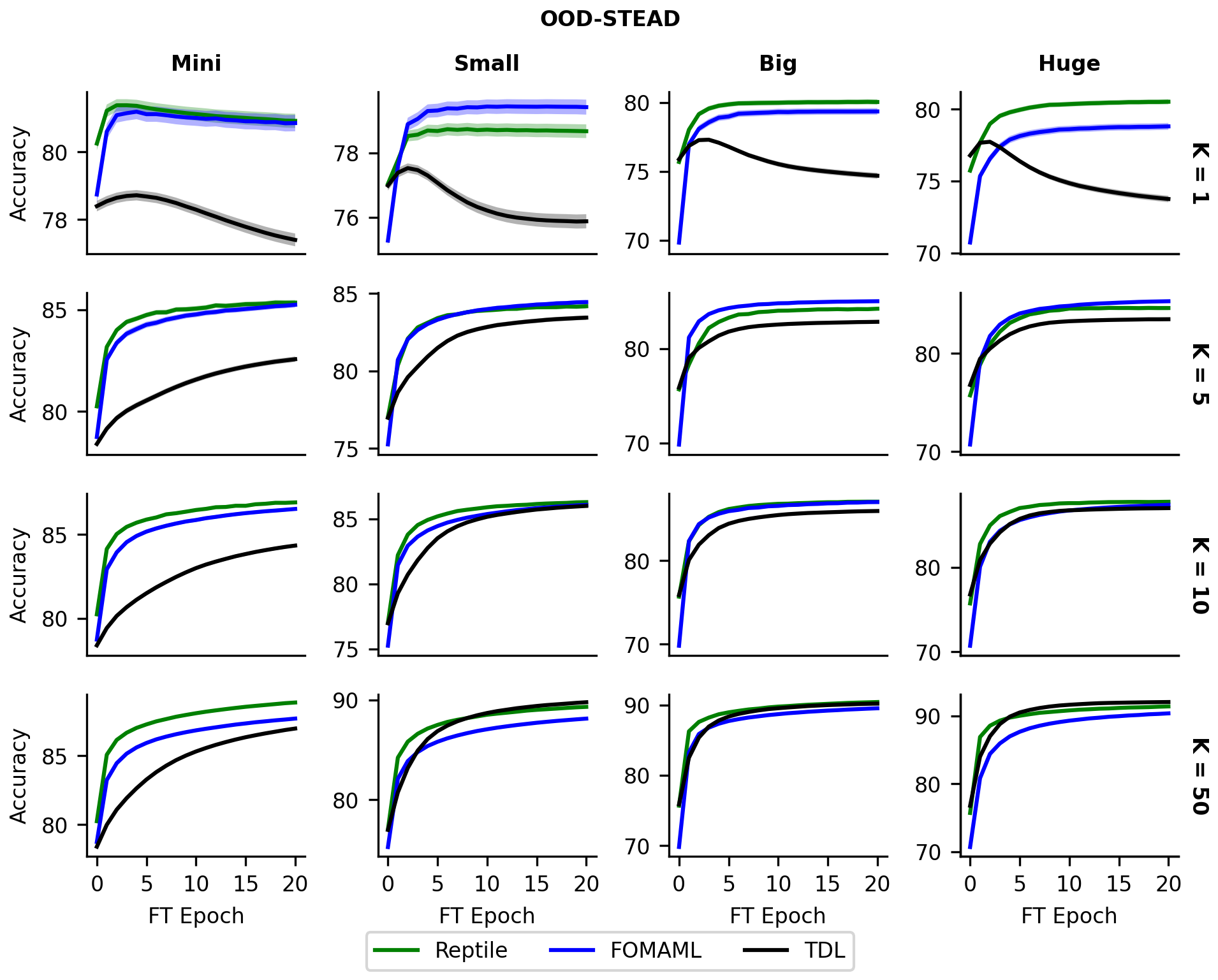}
            \caption{Average performance on \textbf{OOD-STEAD} during fine-tuning, aggregated over training data amount $N$ due to similarity in trends. The x-axis displays the fine-tuning (FT) epoch. Columns are architecture size and rows are the amount of fine-tuning data $K$. For highly shifted OOD-STEAD, meta-learning algorithms adapt faster and better prevent overfitting. As $K$ and architecture size grow, algorithmic performance is similar.}
            \label{fig:loss_STEAD}
        \end{figure}

        We first evaluate how consistently and rapidly each algorithm adapts during fine-tuning, as models that adjust quickly and predictably are best suited for mitigating data shift. The meta-learning algorithms demonstrate faster and more stable adaptation than TDL, particularly when training data are limited or task distributions differ from those seen during training. These trends are illustrated in Figures \ref{fig:loss_SeisTask} and \ref{fig:loss_STEAD}, which show accuracy after each fine-tuning step (FT epoch), and in Figure \ref{fig:loss_speed}, which highlights fine-tuning speed by reporting the average FT epoch at which maximum accuracy is achieved.

        A broad pattern emerges in the fine-tuning curves. The meta-learning algorithms usually exhibit as good or better performance than TDL at earlier epochs (e.g., when FT epoch $\leq5$). If minimal fine-tuning computational resources were available, meta-learning is preferable owing to its quick adaptation. 
        
        However, given enough fine-tuning data, TDL may find better solutions, but it is context-dependent and it is unclear when to stop training without a dedicated validation set. For example, in Figure \ref{fig:loss_SeisTask} with the \textit{big} architecture and $K=1$, TDL eventually reaches a comparatively better solution, but exhibits performance drop-off due to overtraining. While some meta-learning algorithms exhibit overtraining, their consistency in finding strong solutions within few epochs makes them the preferable option. 
    
        Another general trend is that as the architecture size and the amount of fine-tuning data grows, the algorithms tend to converge in terms of convergence speed, stability, and performance, suggesting meta-learning offers more advantages in lower data regimes. 
        
        An interesting phenomenon also pops out when comparing the trends between SeisTask and OOD-STEAD. For SeisTask, the models snap to the optimal solution quickly. For OOD-STEAD, the curves snap to a good solution, but move much slower toward an optimal solution. This indicates the SeisTask test set shares a more similar feature distribution to the training data than OOD-STEAD. 
    
        When comparing the fine-tuning curves between Reptile and FOMAML, there is no obvious winner. While they both exhibit similar speed and consistency, they frequently swap positions in terms of maximum accuracy. For example, in Figure \ref{fig:loss_SeisTask}, Reptile dominates for $K=1$ but is inferior for $K=5$. To complement these trends in adaptation consistency, we next examine overall accuracy across architectures and datasets.
        
    \subsection{Adaptation Performance}
\begin{figure}[h]
            \centering
            \includegraphics[]{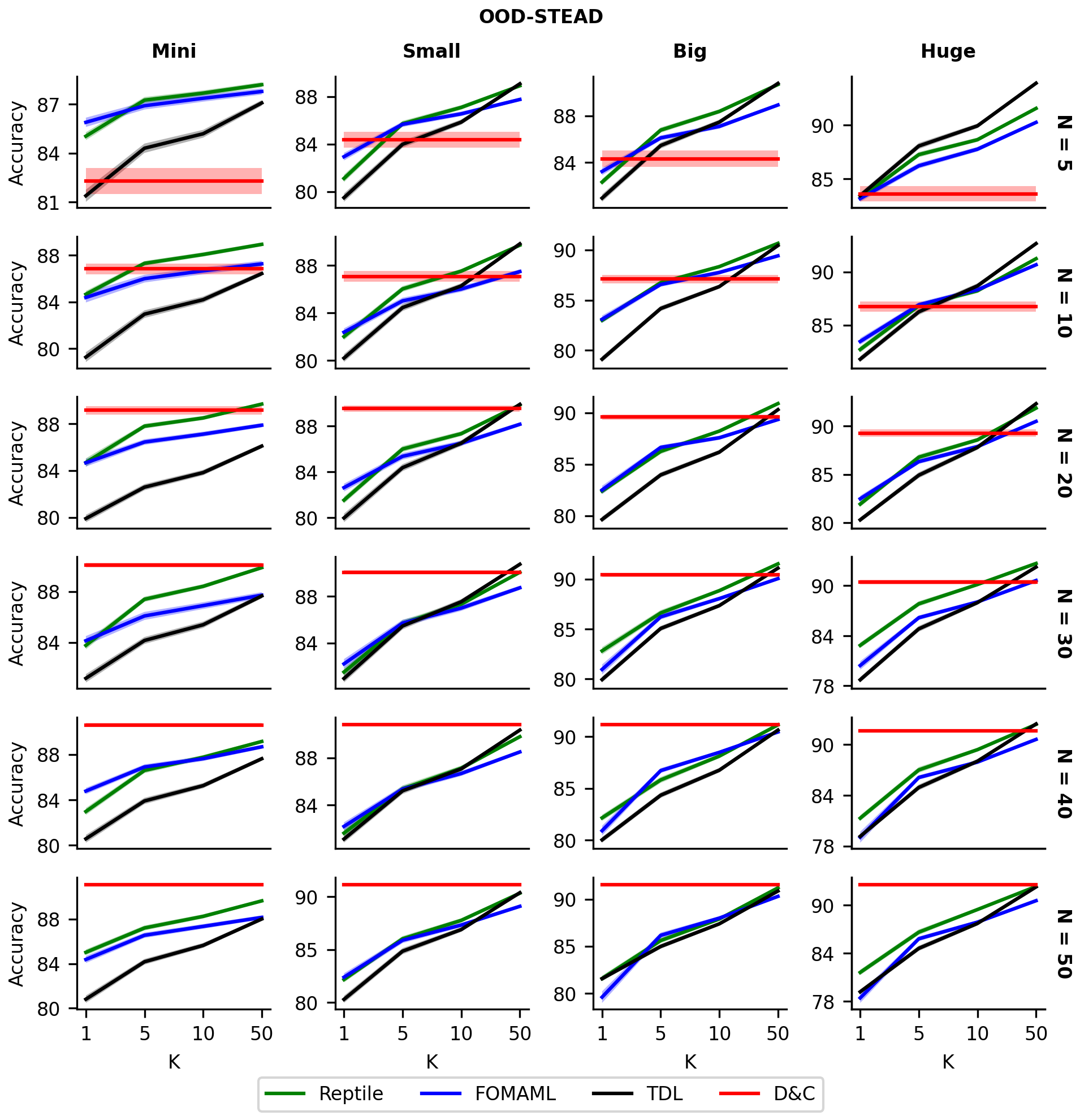}
            \caption{Average accuracy on \textbf{OOD-STEAD} as a function of the amount of fine-tuning data $K$. Columns are architecture size and rows are the amount of training data $N$. Meta-learning algorithms demonstrate strong performance under distribution shift, outperforming TDL particularly when training or fine-tuning data are scarce. However, both approaches fall below the D\&C baseline when $N$ is large, indicating that task-specific training can be advantageous in data-rich regimes.}
            \label{fig:acc_stead}
        \end{figure}

        In evaluating overall adaptation performance, we use the best accuracy achieved throughout fine-tuning to compare each algorithm under ideal conditions. Overall, meta-learning achieves higher or comparable accuracy to TDL across most conditions, with its advantages most evident under strong data shift (OOD-STEAD) or limited training data. We summarize the accuracy scores in Figures \ref{fig:acc_stead} and \ref{fig:acc_seistask}. 
        
        For OOD-STEAD (Figure \ref{fig:acc_stead}), the nearly unanimous better performance of the meta-learning algorithms is convincing. They routinely do as well or better than TDL across most contexts. For SeisTask (Figure \ref{fig:acc_seistask}), relative performance is context-dependent, suggesting interaction effects between data amounts and architecture size. 
    
        When comparing to the D\&C baseline for OOD-STEAD (Figure \ref{fig:acc_stead}), meta-learning and TDL are preferable to D\&C when the amount of training data is small ($N \leq 10$) and there is sufficient fine-tuning data ($K\geq5$). In contrast, for SeisTask, meta-learning and TDL usually outperform the D\&C baseline. These results suggest that more training data results in \textit{dug-in} models that efficiently adapt to test distributions similar to the training distribution but require more fine-tuning data to learn significantly shifted distributions. Accordingly, training from scratch remains a strong alternative when the test distribution is very different from the training distribution. 

        Again, there is no obvious winner between FOMAML and Reptile in terms of overall performance, but Reptile tends to do as well or better than FOMAML in most cases. This could be a product of its fine-tuning procedure where, for each fine-tuning epoch, it benefits from multiple gradient steps compared to a single step in FOMAML. Building on these adaptation results, we next explore whether increasing the diversity of training tasks improves generalization under data shift.

\section{Evaluating Effect of Task Diversity} \label{sec:diversity}
    
    In this section, we examine how the diversity of training tasks influences meta-learning performance by comparing two task-sampling strategies: uniform sampling and diverse sampling. Under uniform sampling, each training task is selected with equal probability during meta-training. Under diverse sampling, tasks are selected with weighted probabilities derived from the task-similarity dendrogram (Figure \ref{fig:dendrogram}), such that each branch of the hierarchy is equally represented, thereby reducing repeated exposure to highly similar tasks (see Appendix \ref{sec:appendix_sampling}).

    Figure \ref{fig:sampling_lines} summarizes the impact of these sampling strategies on performance. We interpret each strategy as inducing a different effective training distribution defined by task sampling probabilities. Using this perspective, we compute the mean similarity between the test tasks and the sampling-induced training distribution, weighted by task selection probabilities, and report these values on the x-axis. For SeisTask (left panel), diverse sampling emphasizes tasks less similar to the test tasks, resulting in degraded performance relative to uniform sampling. In contrast, for OOD-STEAD (right panel), diverse sampling increases exposure to tasks more similar to the test tasks, leading to improved performance. Overall, these results indicate that generalization is driven more by alignment between training and test tasks than by task diversity alone.

    \begin{figure}[h]
        \centering
        \includegraphics[]{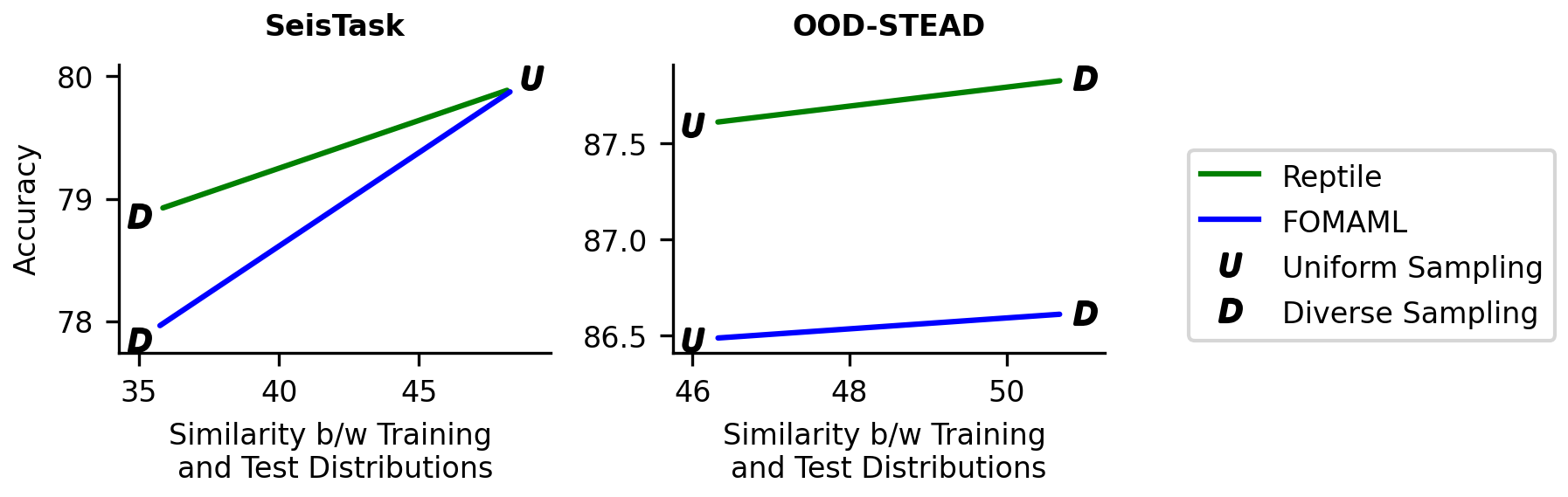}
        \caption{Impacts of emphasizing task diversity. The x-axis corresponds to the mean similarity between the sampling-induced training distribution and the SeisTask test tasks (left) and OOD-STEAD (right). Performance correlates with whether the sampling strategy induces a training distribution that is more similar to the test tasks rather than emphasizing diversity alone. Results are aggregated across all experiments for the \textit{huge} architecture due to similarity in trends.}
        \label{fig:sampling_lines}
    \end{figure}

\section{Discussion \& Conclusion} \label{sec:discussion}
    
    SeisTask represents a valuable dataset that opens numerous avenues for exploration due to its task-oriented structure, strong ground truth, and design-of-experiments construction enabled by its semi-synthetic nature. We demonstrated meaningful similarity relationships among its tasks, establishing SeisTask’s utility within the meta-learning context. It thus serves as a strong benchmark for future time-series meta-learning studies. Future work includes developing general algorithms to identify data characteristics that most strongly influence meta-learning performance and leveraging these insights to inform the design of task-based datasets in other domains.
    
    However, identifying model-relevant data shift in structured datasets remains challenging, particularly when linking such shifts to model performance. Although the similarity measure based on model representations used in this study provides useful insights, it does not fully capture all aspects of distributional shift. For example, OOD-STEAD and SeisTask are moderately similar under this metric, yet fine-tuning convergence on OOD-STEAD is slower (Figure \ref{fig:loss_speed}), suggesting a larger effective shift than indicated by the similarity score. Developing more holistic similarity measures, potentially informed by interpretable deep learning \citep{li_interpretable_2022}, may help better characterize data shift and its impact on model adaptation.
    
    Despite the challenges in precisely characterizing data shift, our results show that meta-learning offers several advantages over TDL for mitigating its effects. Across our experiments, meta-learning algorithms adapt more rapidly and consistently during fine-tuning and achieve better generalization to out-of-distribution tasks, especially for smaller architectures or when data are scarce. However, these advantages are context-dependent: performance varies with the dataset, architecture, and data availability, making it difficult to predict outcomes a priori for new applications.

    In our examination of the role of task diversity in meta-learning, we found that diversity for its own sake did not yield more generalizable models. Instead, generalization performance correlated with whether sampling emphasized tasks more similar to those encountered at test time. We examined this effect using a hierarchical clustering–based task-sampling strategy, though alternative approaches to managing diversity may lead to different outcomes.

    Our study is not without limitations. First, our findings reflect a narrow subset of the broader meta-learning landscape due to our focus on Reptile and FOMAML. The relative performance of these methods varied across settings, and the factors driving these differences remain difficult to disentangle without more targeted experimentation. Additionally, comparative evaluations are inherently influenced by hyperparameter and implementation choices, so it is likely that all methods considered here could benefit from further optimization. Second, we varied training data per task but not the number of training tasks, and performance trends may differ when more or fewer tasks are available. Third, even our largest model is comparatively simple and small relative to many standard AI architectures, reflecting our focus on architectures typical of scientific applications but limiting the scope of architectural conclusions. Extending this work to address these limitations and to other time-series-based problems and domains is a promising direction for future research.

\section{Acknowledgments}
    This research was supported by the U.S. National Nuclear Security Administration (NNSA) Office of Defense Nuclear Nonproliferation Research and Development within the US Department of Energy and by the U.S. Department of Energy through the Los Alamos National Laboratory. Los Alamos National Laboratory is operated by Triad National Security, LLC, for the National Nuclear Security Administration of U.S. Department of Energy (Contract No. 89233218CNA000001). This work is approved for public release under LA-UR-00-0000.


\appendix

\renewcommand\thefigure{\thesection.\arabic{figure}}    

\section{SeisTask Details} \label{sec:appendix_data} \setcounter{figure}{0}
    This section provides significantly more detail regarding the construction of SeisTask, including how the simulator is constructed and data augmentations to prepare SeisTask for use in a deep learning context. 
    
    We built the seismic waveform simulator using the Python package \texttt{Devito} \citep{louboutin_devito_2019, luporini_architecture_2020}. The simulator models subsurface earth structure in 2D and conducts elastic waveform modeling through finite differencing using a system of differential equations to propagate stress and strain across the earth medium. For all simulations we use a square grid of $301\times301$ elements where each elements’ width and height correspond to 10m. We use an additional 10 elements to form an absorbing boundary layer. The simulator requires 3 user-defined items:
    \begin{itemize}
        \item The source describes the initial perturbation of stress that is input into the model at a specific location. 
        \item The earth substructure properties as defined by the compression and shear wave velocities, the speeds that stress travels in parallel and perpendicularly to the propagation of the wave. For simplicity, we define the shear wave velocity as half the compression velocity. 
        \item The receiver is the location at which the 2D waveform signal (velocity) is collected as a function of time.
    \end{itemize}
    
\begin{figure}
        \centering
        \includegraphics[]{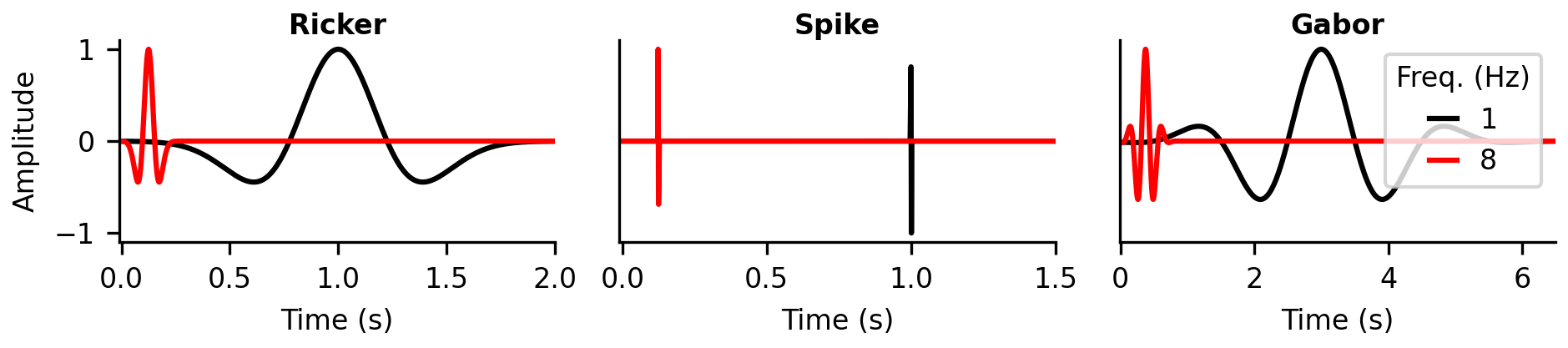}
        \caption{The 3 source shapes: Ricker (left), Spike (middle), and Gabor (right) are visualized at frequencies of 1Hz (black) and 10Hz (red). Higher frequencies for Ricker and Gabor result in a stronger pulse, whereas Spike's width remains constant across frequency and changing frequency is associated with an earlier or later onset.}
        \label{fig:sources}
    \end{figure}

    To create a diverse set of tasks in SeisTask, we implemented a full factorial design. We considered 5 factors (Circles, Layers, Velocity, Frequency, Source) and varied each factor across 3 levels, described as follows.
    \begin{itemize}
        \item Circles defines the number of circles (levels: 0, 2, or 4) in the earth substructure geometry. Each circle is assigned a random radius uniformly between 10m and 50m for each simulator run. 
        \item Layers defines the number of horizontal layer partitions (levels: 0, 2, or 4) in the earth substructure geometry. The partition is randomly assigned at each simulator run using 0 (a homogeneous medium), 2, or 4 slices with a constraint to avoid arbitrarily thin layers of less than 10 m. Layers are generated before the circles, so when circles are added, they overwrite the layer geometry. 
        \item Velocity defines the range of speeds of the compression wave applied to the circles and layers. Each layer and circle are assigned a velocity randomly from the range for each simulator run. We consider ranges across three levels: $\text{Lo}\in[1.50, 3.75]$, $\text{Hi}\in[3.75, 6.00]$, and $\text{Lo} \cup \text{Hi}$. All velocity units are km/s.
        \item Frequency defines the range of frequencies that characterize the width of the source shape. When the simulator is run, the frequency that is randomly pulled from the range. We consider ranges across three levels $\text{Lo}\in[1, 8]$, $\text{Hi}\in[8, 15]$, and $\text{Lo} \cup \text{Hi}$. All frequency units are in Hz.
        \item Source defines the wavelet shape of the event. We consider three types: Ricker, Spike, and Gabor. Examples of each source at two different frequencies are shown in Figure \ref{fig:sources}.
    \end{itemize}
    
\begin{figure}
        \centering
        \includegraphics[]{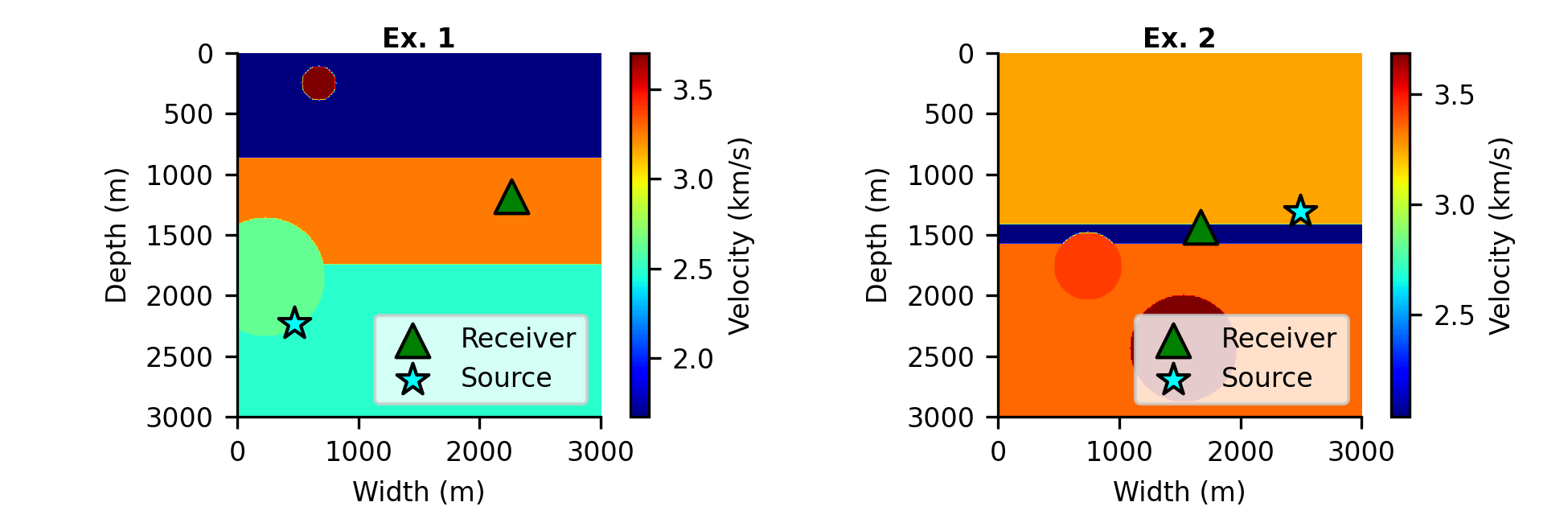}
        \caption{Two examples of the earth substructure and source and receiver locations for tasks associated with 2 circles and 2 layer partitions are shown. Each run, or replicate, of the seismic simulator produces a random earth substructure like the examples shown.}
        \label{fig:profiles}
    \end{figure}

    At each unique combination of factor levels, we run the simulator 210 times to obtain 210 event signatures. Each of the 210 event signatures are unique due to the random elements associated with the fixed factor levels and because we randomly assign the source and receiver locations each simulator run (requiring the source to receiver distance to be a minimum of 30m). In addition, we minimize boundary layer interactions by restricting the source and receiver locations to be at least 50m from the boundary layer. Figure \ref{fig:profiles} shows two simulator runs (i.e., examples or replicates) that contain 2 circles and 2 layer partitions to demonstrate how each simulator run produces random substructure geometries, source and receiver locations, and velocities at fixed factor levels. The result is $3^5=243$ unique groups of event signatures arising from the design that each contain 210 unique event signatures. Finally, we constrain the simulator to produce waveforms sampled at 100Hz. 

    After simulating 210 pure (noiseless) signals for each setting in the design, we curate the data for the classification model. For each pure signal, we trim it to 5s and randomly shift each window such that the initial event signature (the compression wave) arrives randomly between 0.50s and 4.50s. This ensures the model classifier cannot systematically memorize that a signal occurs at the same time for every waveform. 

    To make the effort of classification more challenging and realistic, we add real noise to each pure signal from the \texttt{SeisBench} \citep{woollam_seisbenchtoolbox_2022} implementation of INSTANCE according to the following process. We first calculate the power of each component in the pure signal and scale each component by the largest of the two powers. This ensures that the maximum power of any component is unity, while maintaining the relative power between each component. Then, we sample a random noise waveform from INSTANCE and normalize it in the same way but conduct a final scaling such that the maximum power is 5. We then add the scaled pure signal to the scaled noise and re-normalize the result to have a maximum absolute value of 1 to allow for easier training convergence. This process ensures we maintain a consistent signal-to-noise ratio (SNR) of $1/5$ across all signal waveforms. We refer to the resulting signals that are embedded in noise as signal waveforms (Figure \ref{fig:wf_examples}). 

    To form the tasks in SeisTask, we curate noise-only waveforms to complement the signal waveforms in each treatment. To do so, we randomly select 243 disjoint groups of 210 noise waveforms from INSTANCE. Then, we randomly window and trim each waveform to 5s and normalize such that the maximum absolute value is 1. All noise samples are taken without replacement, so they are all unique.
    
    Finally, we define a task by pairing each treatment’s 210 signal waveforms with a randomly selected group of 210 noise waveforms, thereby ensuring class balance for each task. The end result is SeisTask, composed of 243 tasks each containing 420 waveforms of 210 signal waveforms and 210 noise waveforms. For OOD-STEAD, to prepare the data for modeling, we randomly trim, window, and normalize the waveforms consistent with SeisTask such that the signals arrive between 0.50s and 4.50s, all waveforms are 5s long in 2D (we drop the East-West dimension), and the maximum absolute value is 1. In the sections that follow, we represent SeisTask as $\mathcal{T}=\{\tau_t\}_{t=1}^{T=243}$ and OOD-STEAD as $\mathcal{T}^*=\{\tau_t^*\}_{t=1}^{T^*=35}$. 

\section{Deep Learning Architecture Details} \label{sec:appendix_models} \setcounter{figure}{0}
    This section provides more details about the deep learning architectures. Motivated the demonstrated success of convolutional neural networks (CNNs) in seismic time series \citep{zhu_phasenet_2018}, we designed an original base architecture tailored for classification that is easily scalable to different sizes. This base architecture consists of a one-dimensional (1D) CNN feature extractor, a global average pooling layer, a fully connected multi-layer perceptron (MLP) classifier, and a final sigmoid activation for binary prediction. Rectified linear unit (ReLU) activation functions are used between each layer. Its input is a 2D waveform $x_i \in \mathbb{R}^{2 \times 500}$ and its scalar output $\hat{y}_i \in [0,1]$ represents the probability that the waveform contains a signal ($\hat{y}_i \geq 0.5$) or is pure noise ($\hat{y}_i < 0.5$). Binary cross entropy is used as the loss function
    
    To explore how architecture size impacts performance, we scale the base architecture across four variants (\textit{mini}, \textit{small}, \textit{big}, and \textit{huge}) where that increase in model capacity by deepening and widening both the CNN and MLP components. The four architectures differentiate themselves as follows:
    \begin{itemize}
        \item \textbf{Mini} (389 trainable parameters), the smallest configuration, serves as a lightweight baseline. Its CNN stack contains 2 layers with channel counts of 4 and 8 filters with a max-pooling operation. The MLP stack consists of layers of sizes 8 and 16.
        \item \textbf{Small} (3,921 trainable parameters) builds upon \textit{mini} by widening the CNN stack to 16 and 32 filters. The MLP head is also widened to sizes 32 and 64.
        \item \textbf{Big} (24,593 trainable parameters) deepens the CNN stack to three layers with channel counts 16, 32, and 64. The MLP stack is deepened to three layers of sizes 64, 128, and 64.
        \item \textbf{Huge} (274,017 trainable parameters) deepens the CNN stack to four layers with channel counts 32, 64, 128, and 192 with slightly larger kernel sizes and includes two max-pooling operation operations. Its MLP stack widens to sizes 192, 384, and 192.
    \end{itemize}

\section{Data Shift in SeisTask Details} \label{sec:appendix_datashift}
    This section provides more details about how data shift in SeisTask is quantified. All task-specific models in this section used the largest network (\textit{huge}) and were trained with $N=50$ (the number of training samples per class and per task, which provided the clearest representation of task relationships). To mitigate variability due to the stochastic training process, we trained 10 independent ensemble task-specific models for each task using only that task’s data, without exposure to any other tasks. 
    
    \subsection{Cross-task Accuracy}
        This section aims to formalize the description of cross-task accuracy provided in the main text. We construct a crosswise accuracy matrix $A \in \mathbb{R}^{T \times T}$ containing elements $a_{uv}$ representing the accuracy of a task-specific model trained on task $u$ and evaluated on task $v$ (averaged over the $10$ ensembles). We analyze $A$ to determine whether task-specific models learn features that are specific to each task or transferable across tasks. To summarize these relationships, we compute the proportion $p_u$ for each evaluation task $u$ as
        \begin{equation}
        p_u = \frac{1}{T-1} \sum_{v=1, v \ne u}^T \mathbb{I}\{a_{uu} > a_{vu}\},
        \label{eq:pu}
        \end{equation}
        where $p_u \in [0,1]$ and $\mathbb{I}(\cdot)$ is the indicator function.

    \subsection{Similarity}
        To compute similarity using the model representations, we leverage linear Center Kernel Alignment to compute similarity between tasks. Allow $X\in \mathbb{R}^{r \times c}$ and $Y \in \mathbb{R}^{r \times c}$ be column-centered matrices of activations of the same $r$ observations in $q$ dimensions, then the linear Center Kernel Alignment similarity metric is given as 
        \[
        s(X,Y)=100\frac{|Y^T X|^2_F}{|X^T X|^2_F |Y^T Y|^2_F}
        \]
        where $|\cdot|_F$ indicates the Frobenius norm and $s(X,Y)\in[0,100]$. Larger values of $s(X,Y)$ indicate high similarity as viewed by the model between $X$ and $Y$. The scaling by $100$ is arbitrary to allow for easier digestion. 

        We compute the similarity between tasks $u$ and $v$ as follows. Allow $U(\mathcal{D})$ and $V(\mathcal{D})$ to represent the activations of data $\mathcal{D}$ arising from the model trained on task $u$ and $v$, respectively. In our case, we collect the data-activations after the global average pooling operation between the CNN and MLP stacks. Allow $\tau_{uv} = \begin{bmatrix} \tau_u \\ \tau_v \end{bmatrix}$ be the concatenated data for each task. We construct a symmetric similarity matrix $S \in \mathbb{R^{T \times T}}$ with elements $s_{uv} = s(U(\tau_{uv}), V(\tau_{uv}))$. We average over the ensembles for each $U(\cdot)$ and $V(\cdot)$ ($10^2$ pairs for non-diagonal elements and $\binom{10}{2}$ for the diagonal elements). 

        The plot for Figure \ref{fig:sim_acc} is described as follows. We first compute a symmetric form of $A$ as $A' = A^T A / 2$, whose elements reflect the expected accuracy when models are trained on one task and evaluated on another. Plotted are the upper triangular matrices of $A'$ (y-axis) and $S$ (x-axis), excluding the diagonals, where each point represents paired tasks binned by similarity.
        
    \subsection{Agglomerative Clustering for Task Splits}
\begin{figure}[H]
            \centering
            \includegraphics[]{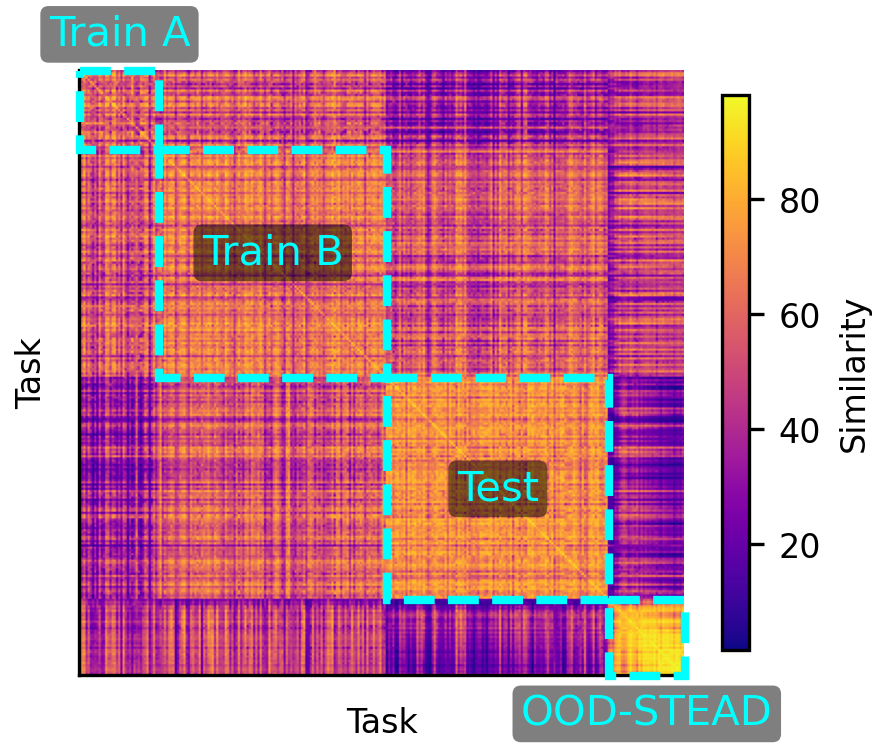}
            \caption{The organized similarity matrix $S$ according to the assignments visually shows intra-group similarity and inter-group dissimilarity, demonstrating data shift between the training and validation tasks (Train A and B) and the others (Test and STEAD).}
            \label{fig:sim_mat_organized}
        \end{figure}

\begin{figure}[h]
            \centering
            \includegraphics[]{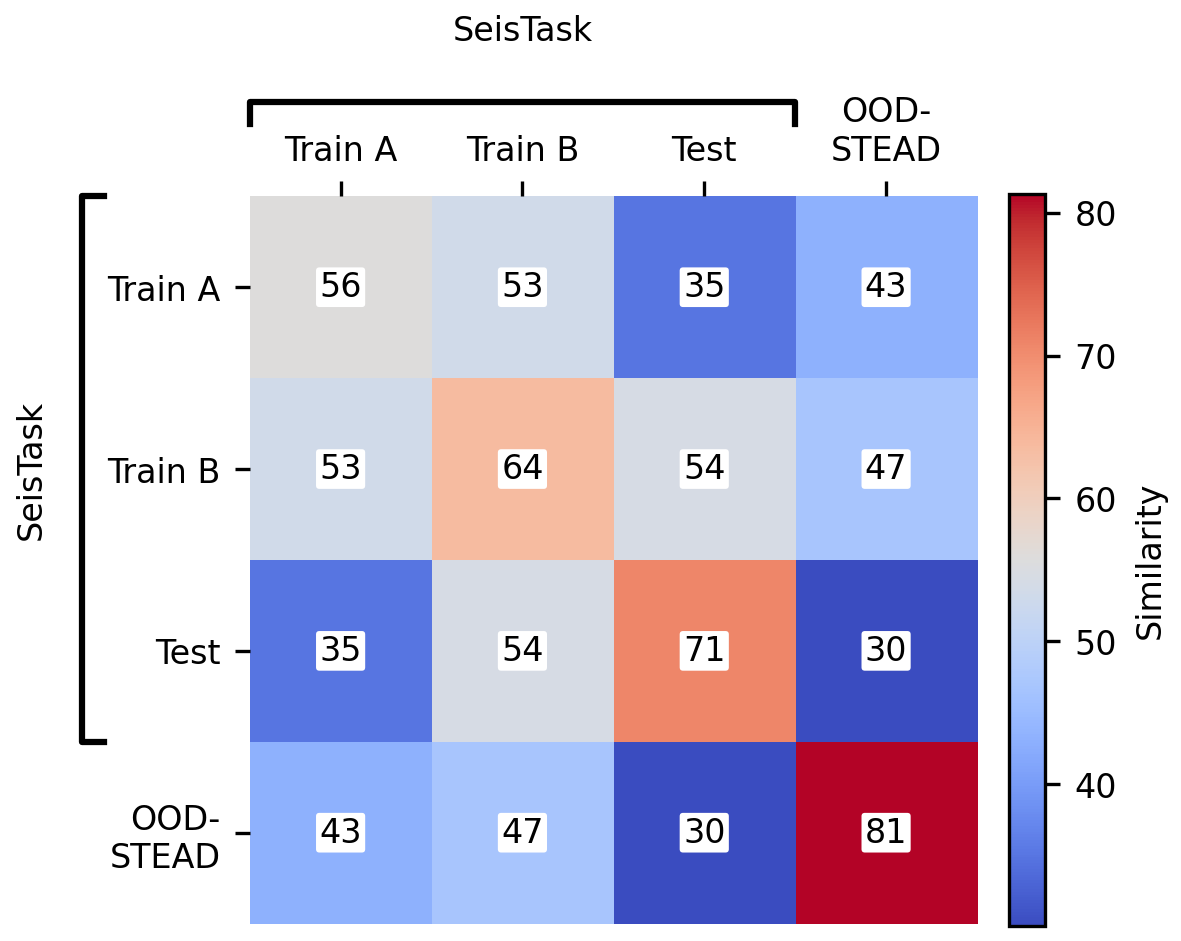}
            \caption{Reduced heatmap of the average similarities between each subgroup under uniform sampling. The SeisTask test group is less similar to Train A than to Train B. Diverse sampling places greater emphasis on Train A than uniform sampling.}
            \label{fig:sampling_reduced_heatmap}
        \end{figure}

        From the similarity matrix $S$, we form the distance matrix $D$ with elements $d_{uv}$ as 
        \[
        d_{uv}=
        \begin{cases}
          100-s_{uv} & \text{if } u \neq v \\
          0  & \text{if } u=v\text{.}
        \end{cases}
        \]
        The criteria setting the diagonal elements to $0$ is needed because $s_{uu}$ is not necessarily $100$ due to variability in the ensemble training. We use $D$ to construct the dendrogram in Figure \ref{fig:dendrogram}. In future sections, we denote the SeisTask test tasks as $\mathcal{T}^{Te}$ and the tasks in the training and validation pool as $\mathcal{T}^{Pool}$. An organized similarity matrix showing the relationships between the SeisTask test tasks, training and validation pool (Train A and Train B), and OOD-STEAD tasks (included as supplementary information), is shown in Figure \ref{fig:sim_mat_organized} and is summarized in Figure \ref{fig:sampling_reduced_heatmap}. 
        
    \subsection{Uniform and Diverse Sampling Details} \label{sec:appendix_sampling}
        To emphasize diversity, we employ two sampling strategies: diverse sampling and uniform sampling. The sampling strategies are distinguished by how the task weights $\gamma_t$'s are constructed. The task weights influence the construction of the training and validation splits prior to training each model as well as the probably of each task being selected during meta-training. The $\gamma_t$ values only correspond to the tasks in the training and validation pool. 
        
        Under the uniform sampling strategy, all tasks have equal probability of being selected, such that $\gamma_t $ is the same for all $t$. The validation split $\mathcal{T}^{Val}$ is created by randomly sampling $20\%$ of the tasks in $\mathcal{T}^{Pool}$, while the remaining tasks form the training split $\mathcal{T}^{Tr}$. During meta-training, tasks from $\mathcal{T}^{Tr}$ are selected uniformly without replacement. 
        
        In contrast, the diverse sampling strategy biases selection toward tasks that are less similar on average, thereby encouraging exposure to a wider range of task distributions. To compute the diversity weights, we construct a new dendrogram but only using tasks in $\mathcal{T}^{Pool}$. Then, each task’s weight is defined as $\gamma_t = 2^{-k_t}$, where $k_t$ is the number of bifurcations from the dendrogram root to the task’s leaf node. Validation tasks $\mathcal{T}^{Val}$ are chosen iteratively based on $\gamma_t$, where the dendrogram is rebuilt to update the weights after each selection to reduce bias toward dense clusters near the root. The tasks not selected for the validation split become the training split $\mathcal{T}^{Tr}$. During meta-training, tasks from $\mathcal{T}^{Tr}$ are sampled without replacement according to their updated diversity weight $\gamma_t$ values, as described in Algorithms \ref{alg:reptile} and \ref{alg:fomaml}.
    
\section{Algorithmic Details} \label{sec:appendix_algs} \setcounter{figure}{0}

    \subsection{Training}
        This section provides significantly more details for how Reptile, FOMAML, TDL, and D\&C were implemented, including more background, training procedures, hyperparameter selection, and detailed algorithms. 
        
        The first-order variants FOMAML and Reptile approximate MAML while avoiding its computationally expensive second-order derivatives. Both aim to find flexible parameters $\phi$ that can be quickly adapted to $\theta_t$ using few samples per class (few-shot learning). First, the data in each task are relegated to either the support or query partition. Then, Each iteration proceeds in two stages. The inner stage adapts $\phi$ to task-specific parameters $\theta_t$ using the support partition. The outer stage updates $\phi$ based on how well $\theta_t$ performs on the query partition. 
        
        The main differences between FOMAML and Reptile are in the implementation of how data are used during training and in implementation of the meta-update. For the data-utilization on the inner-step, FOMAML fine-tunes using the task's support partition and conducts one SGD step, whereas Reptile uses support \textit{and} query data and incrementally fine-tunes using $G$ SGD steps. For the meta-update, FOMAML utilizes a gradient over gradients in the outer-loop for the query set, whereas Reptile updates the meta-model by moving from $\phi$ towards $\theta_t$ incrementally (for example, through the linear operation $\phi \leftarrow \phi + \beta (\theta_t - \phi)$, where $\beta$ is the outer loop training rate).

        Here, we describe how we systematically vary the amount of training data $N\in\{5, 10, 20, 30, 40, 50\}$ in SeisTask through task partitions. For a given $N$, each task is randomly partitioned into disjoint support, query, and k-shot partitions, each containing $N$ pairs per class, with the remaining data assigned to the hold-out partition. Generally, the support and query partitions are used for training, the k-shot partition for fine-tuning, and the hold-out partition for evaluation, however the exact use differs slightly by algorithm as describe below. In the algorithms that follow, we refer to the support, query, k-shot, and hold-out partitions formally as $\tau_t^{Sup}$, $\tau_t^{Que}$, $\tau_t^{K}$, and $\tau_t^{Hold}$, respectively.
        
        Since OOD-STEAD is used exclusively for evaluation, we partition each task into a hold-out partition and a k-shot partition. The k-shot partition contains $N$ pairs per class and the remaining waveforms are assigned to the hold-out partition. 

        We first consider Reptile, which trains a model optimized for efficient learning in $N$ data samples (per class and per task) through $G$ SGD steps with learning rate $\alpha$ (SGD-$\alpha$). To constrain the hyperparameter space over which we optimize, we consider $G=5$ for all experiments. For the inner loop, we sample $5$ training tasks and, in parallel, train from $\phi$ to task-specific parameters $\theta_t$ using randomly selecting $N$ pairs per class from the task's support and query partitions. For the meta-update, we compute the average of the $5$ $\theta_t$'s, denoted $\bar{\theta}$ and use $(\phi - \bar{\theta})$ as the gradient for a single ADAM-$\beta$ step. For the validation step, we perform $N$-shot fine-tuning using each validation task’s support partition and compute the loss on its query partition. Reptile’s optimization procedure is detailed in pseudo-code in Algorithm \ref{alg:reptile}. 
        
        \begin{algorithm}
        \caption{Reptile Optimization Procedure}
        \label{alg:reptile}
        \begin{algorithmic}
            \State Randomly split $\mathcal{T}$ into disjoint $\mathcal{T}^{Val}$ and $\mathcal{T}^{Tr}$ sets with selection probabilities $\gamma_t$
            \State Randomly initialize $\phi$
            \For{each epoch}
                \For{each pseudo-epoch} \Comment{Train}
                    \State Sample $5$ tasks from $\mathcal{T}^{Tr}$ without replacement with selection probabilities $\gamma_t$ to make $\mathcal{T}^{Tr}_5$
                    \For{$\tau_t \in \mathcal{T}^{Tr}_5$} \Comment{Task Specific Updates}
                        \State Randomly sample $N$ samples per class from $\{\tau_t^{Sup}, \tau_t^{Que} \}$ to create $\mathcal{D}_t$ 
                        \State Train on $\mathcal{D}_t$ in 5 SGD-$\alpha$ steps starting with $\phi$ resulting in parameters $\theta_t$ 
                    \EndFor
                    \State Compute average of tuned weights $\bar{\theta}=\frac{1}{5} \sum_{t=1}^5\theta_t$
                    \State Update $\phi$ via ADAM-$\beta$ using $(\phi-\bar{\theta})$ as the gradient \Comment{Meta Update}
                \EndFor
                
                \For{$\tau_t \in \mathcal{T}^{Val}$} \Comment{Validate}
                    \State Train on $\tau_t^{Sup}$ in 5 SGD-$\alpha$ step starting with $\phi$ resulting in parameters $\theta_t$
                    \State Compute loss: $l^{Val}_t = \mathcal{L}(\theta_t, \tau_t^{Que}) (2N)^{-1}$
                \EndFor
                \State Compute average validation loss: $|\mathcal{T}^{Val}|^{-1} \sum_{t=1}^{|\mathcal{T}^{Val}|} l^{Val}_t$
        
            \EndFor
        \end{algorithmic}
        \end{algorithm}
        
        For FOMAML, the training process is similar in structure as Reptile, but diverges in terms of the gradient update and batching. On the inner loop, we sample $5$ tasks, training in 1 SGD-$\alpha$ step using $N$ pairs per class sampled form the support and query partitions and compute the task-loss $l_t$. For the meta-update, we compute the average task-loss $\bar{l}$ and update $\phi$ via a single ADAM-$\beta$ step using the gradient of $\bar{l}$ with respect to $\phi$. For validation, we fine-tune on each validation task using the support partition and compute the loss on the query partition. FOMAML's optimization procedure is detailed in pseudo-code in Algorithm \ref{alg:fomaml}. 
        \begin{algorithm}
        \caption{FOMAML Optimization Procedure}
        \label{alg:fomaml}
        \begin{algorithmic}
            \State Randomly split $\mathcal{T}$ into disjoint $\mathcal{T}^{Val}$ and $\mathcal{T}^{Tr}$ sets with selection probabilities $\gamma_t$
            \State Randomly initialize $\phi$
            \For{each epoch}
                \For{each pseudo-epoch} \Comment{Train}
                    \State Sample $5$ tasks from $\mathcal{T}^{Tr}$ without replacement with selection probabilities $\gamma_t$ to make $\mathcal{T}^{Tr}_5$
                    \For{$\tau_t \in \mathcal{T}^{Tr}_5$} \Comment{Task Specific Updates}
                        \State Sample disjoint datasets $\mathcal{D}^{Sup}_t$ and $\mathcal{D}^{Que}_t$ of $N$ samples per class from $\{\tau_t^{Sup}, \tau_t^{Que} \}$
                        \State Train on $\mathcal{D}^{Sup}_t$ in 1 SGD-$\alpha$ step starting with $\phi$ resulting in parameters $\theta_t$
                        \State Compute $l_t = \mathcal{L}(\theta_t, \mathcal{D}^{Que}_t) $ 
                    \EndFor
                    \State Compute meta-loss $\bar{l} = \frac{1}{5}\sum_{t=1}^5 l_t$
                    \State Update $\phi$ via ADAM-$\beta$ using $\nabla_\phi \bar{l}$ as the gradient \Comment{Meta Update}
                \EndFor
        
                \For{$\tau_t \in \mathcal{T}^{Val}$} \Comment{Validate}
                    \State Train on $\tau_t^{Sup}$ in 1 SGD-$\alpha$ step starting with $\phi$ resulting in parameters $\theta_t$
                    \State Compute loss: $l^{Val}_t = \mathcal{L}(\theta_t, \tau_t^{Que})(2N)^{-1}$
                \EndFor
                \State Compute average validation loss: $|\mathcal{T}^{Val}|^{-1} \sum_{t=1}^{|\mathcal{T}^{Val}|} l^{Val}_t$
            \EndFor
        \end{algorithmic}
        \end{algorithm}
        
        For TDL, which trains in a task-agnostic manner, we first combine the training and validation tasks together. Then, to keep the training and validation sizes approximately equal to Reptile, we randomly select 80\% of each tasks’ support and query partitions to form the training data and the remaining 20\% to form the validation data while maintaining class balance in each set. This results in approximately equal training and validation sizes between TDL and meta-learning, allowing for more fair comparison. TDL’s optimization procedure is detailed in pseudo-code in Algorithm \ref{alg:tdl}. The SGD training batch size for Reptile is given as $BS_{Rept}=N/G$. For TDL, we fixed the batch size to $16$. 
        \begin{algorithm}
        \caption{TDL Optimization Procedure}
        \label{alg:tdl}
        \begin{algorithmic}
            \State Pool data across tasks and partitions: $\mathcal{D}=\{ \tau_t^{Sup} \cup \tau_t^{Que} \} \forall \tau_t \in \mathcal{T}$
            \State Form disjoint sets $\mathcal{D}^{Tr}$ and $\mathcal{D}^{Val}$ via an 80-20 split of $\mathcal{D}$ with class balance
            \State Randomly initialize $\phi$
            \For{each epoch}
                \State Train on $\mathcal{D}^{Tr}$ in $\lceil \frac{|D^{Tr}|}{16} \rceil$ ADAM-$\beta$ steps updating $\phi$ \Comment{Train}
                \State Compute average validation loss: $ |\mathcal{D}^{Val}|^{-1} \mathcal{L}(\phi,\mathcal{D}^{Val})$ \Comment{Validate}
            \EndFor
        \end{algorithmic}
        \end{algorithm}
    
        For D\&C, for a given task, we train a model using $3N$ pairs per class with an 80/20 random train/validation split, repeating this process 10 times under different random initializations to reduce variability. We repeat this procedure across all tasks, architectures, and training data amounts $N$. D\&C's optimization procedure is detailed in pseudo-code in Algorithm \ref{alg:dnc}.
        \begin{algorithm}
        \caption{Divide and Conquer (D\&C) Optimization Procedure}
        \label{alg:dnc}
            \begin{algorithmic}
                \State Given $\tau_t \in \mathcal{T}$
                \State Make $\mathcal{D}_t=\{\tau_t^{Sup} \cup \tau_t^{Que} \cup \tau_t^{K-shot} \}$
                \State Form disjoint sets $\mathcal{D}_t^{Tr}$ and $\mathcal{D}_t^{Val}$ via an 80-20 split of $\mathcal{D}_t$ with class balance
                \State Randomly initialize $\phi$
                \For{each epoch}
                    \State Train on $\mathcal{D}_t^{Tr}$ in $\lceil \frac{|D_t^{Tr}|}{16} \rceil$ ADAM-$\alpha$ steps updating $\phi$ \Comment{Train}
                    \State Compute average validation loss: $| \mathcal{D}^{Val}_t | \mathcal{L}(\phi, \mathcal{D}^{Val}_t)$ \Comment{Validate}
                \EndFor
            \end{algorithmic}
        \end{algorithm}
            
        In TDL parlance, an \textit{epoch} is defined as one full loop through the training data during training. However, this definition is less applicable for meta-learning, which typically trains in \textit{iterations}. To reconcile these language gaps, we define pseudo-epochs as the number of iterations needed to achieve approximately one lap through the training data. For example, we use task-batches of size 5, so the number of pseudo-epochs is $\lceil \frac{T}{5} \rceil$.
        
        We conducted a series of preliminary tests to determine the optimal hyperparameters for each of the 72 unique combinations of modeling approach (Reptile, FOMAML, and TDL), architecture size, and $N$. For these models, we varied $\beta \in \{ 1e^{-1},1e^{-2},1e^{-3}, 5e^{-4}, 1e^{-4} \}$ and, for the meta-learning approaches, we also varied the inner learning rate $\alpha \in \{ 1e^{-1},1e^{-2},1e^{-3}, 1e^{-4} \}$. We found that models with hyperparameter settings of $\beta=5e^{-4}$ and $\alpha=1e^{-2}$ trained in the most stable manner and produced the best performing models on the validation data, so we used these settings for all final models. Training continued until the mean validation loss failed to improve for 200 consecutive epochs. 
        
        For the D\&C models, we used a learning rate $\beta$ of $5e^{-4}$ and a batch size of $16$ based on similar preliminary experiments. Training continued until the mean validation loss failed to improve for 150 consecutive epochs.       
    
        In the above training approaches, it may not be evident that each algorithm using approximately the same amount of data. However, we make a concerted effort to ensure this is the case as described as follows. The meta-learning approaches use 112 training tasks' support and query partitions ($112 \times 2N$) and validates using the 29 validation tasks’ support and query waveforms ($29 \times 2N$). TDL uses the query and support partitions from all 141 tasks in $\mathcal{T}^{Pool}$, but with an 80-20 split into training ($141 \times 0.8 \times 2N \approx 112*2N$) and validation ($141 \times 0.2 \times 2N \approx 28 \times 2N$). 

    \subsection{Fine-tuning} 
        This section describes more details about how each model is fine-tuned on each task. 
        
        To evaluate how efficiently each model adapts to new tasks, we fine-tuned models using $K$ samples per class drawn from each task's k-shot partition. We examined performance across $K \in \{1, 5, 10, 50\}$ to capture a range from limited to relatively rich data regimes. Fine-tuning was performed for $\{0, 1, \ldots, 20\}$ gradient steps, where $0$ indicates evaluation without fine-tuning. After each step, model performance was assessed on the corresponding task's hold-out partition. In contrast to the meta-learning algorithms, which are optimized for fine-tuning, the TDL approach is not natively geared for fine-tuning. However, for fair comparison, we provide TDL the same opportunity to fine-tune. For the D\&C models, no fine-tuning is conducted, as each task-specific model was already trained on the test task on which it is evaluated. 
        
        In general, for the meta-learning algorithms, the fine-tuning is conducted as in the validation step of their respective algorithms. An exception to this is for the edge case in Reptile when $K<5$, since we cannot conduct $5$ class-balanced gradient steps without repeating over the same data when $K<5$. To reconcile this issue, we set lower the number of gradient steps to the fine-tuning data size (e.g. $G=K$). For TDL, we adapt the inner batch size to account for the generally smaller $K$ by requiring $5$ gradient steps like the Reptile algorithm. 
        
        At each gradient step, we pause fine-tuning to conduct evaluation. For SeisTask, since the tasks in $\mathcal{T}^{Te}$ were not used during training regardless of $N$ (except for when training the D\&C models), we pool the data from the support, query, and hold-out partitions to be the evaluation data (totaling $160$ pairs per class), the remaining $50$ pairs per class are sampled randomly depending on $K$ for fine-tuning. For OOD-STEAD, we sample $K$ waveforms from the k-shot partition and use the hold-out partition for evaluation ($250$ pairs per class). The models are always fine-tuned with the same $K$ waveforms across approaches and evaluated against the same data for proper comparison.

\section{Evaluation and Uncertainty Quantification Details} \label{sec:appendix_uncertainty}\setcounter{figure}{0}
        This section outlines the evaluation metrics in greater detail and provides arguments supporting our uncertainty quantification approach for constructing error bars. Excluding the D\&C models, the total result from these experiments yields 2,400 independently trained models from three modeling approaches, four architecture sizes, six training data amounts ($N$), two task-sampling strategies (TDL employs only uniform sampling due to its task-agnostic training), and 20 ensembles per configuration, all of which are evaluated for each fine-tuning data amount $K$.
        
        To summarize this information, we compare model performance based on classification accuracy. Given the algorithm, architecture, $N$, and $K$, we compute the accuracy $Acc_{te}$ on task $t$ from ensemble ($e \in \{1, \dots, E=20\}$) in any task-split $\mathcal{T}^{\cdot}$ containing $T^\cdot$ tasks. We compute the mean accuracy for the task across the ensembles as $Acc_t = \sum_{e=1}^{E} Acc_{te} / E$. Then, we compute the average accuracy over the tasks to summarize performance over the split as $\overline{Acc}  = \sum_{t=1}^{T^\cdot} Acc_t / {T^\cdot}$. When trends were consistent across multiple values of $N$ or $K$, results were further aggregated as indicated in the corresponding figure captions. 

        To account for uncertainty in our results, we model $Acc_{te} \sim \mathcal{N}(\alpha_t, \sigma^2_{t})$, where $\alpha_t$ is the quantity of interest estimated using $Acc_t$ (the sample mean) and $\sigma^2_t$ (the sample variance). Here, $\mathcal{N}(\mu, v^2)$ represents the Normal distribution with mean $\mu$ and variance $v^2$. This imbues a distribution on $Acc_t$ as $Acc_t \sim \mathcal{N}(\alpha_t, \sigma^2_t / E) $. When aggregating via the mean over tasks in a split computing $\overline{Acc}$, we get $\overline{Acc} \sim \mathcal{N}(\alpha, \sigma^2)$ where $\alpha={T^{\cdot}}^{-1} \sum_{t=1}^{T^\cdot} \alpha_t$ and $\sigma^2={T^\cdot}^{-2} \sum_{t=1}^{T^\cdot} \sigma^2_t/E$. Error bars/ribbons in the figures represent 95\% confidence intervals according to these assumed distributions. 
        
\begin{figure}[H]
            \centering
            \includegraphics[]{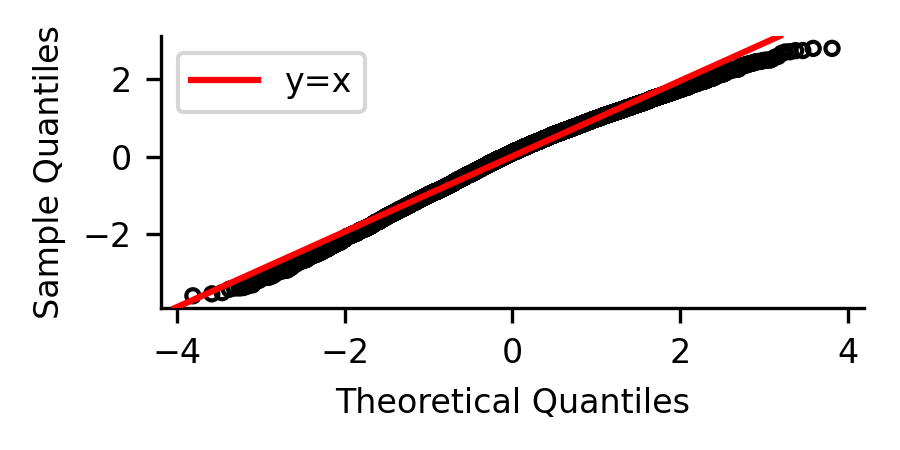}
            \caption{QQ plot of sample quantiles versus the theoretical quantiles for a subset of 10,000 residuals ($Acc_{te} - Acc_t$) computed across modeling approaches, training data amounts, fine-tuning data amounts, architecture sizes, and sampling strategies. The trend follows closely to the $y=x$ line (red), suggesting moderate-to-strong confirmation of our statistical assumptions used to quantify error.}
            \label{fig:qq}
        \end{figure}

        To verify our distributional assumptions, we provide a QQ-plot (Figure \ref{fig:qq}) comparing a subset of the 10,000 empirical residuals of $Acc_{te} - Acc_t$ collected across all 2,400 models. In Figure \ref{fig:qq}, we see fairly strong support in our assumptions, where the sample quantiles and theoretical quantiles follow the $y=x$ line. There is some evidence of mildly heavier tails than assumed, but our goal is to provide statistically informed uncertainty, so these deviations are minor in context of the purpose of our study.
        
\section{More Results} \label{sec:appendix_results} \setcounter{figure}{0}
    The following figures are referred to in the main text, but their slightly redundant information merits their inclusion in the appendix.
    
    \begin{figure}[H]
        \centering
        \includegraphics[]{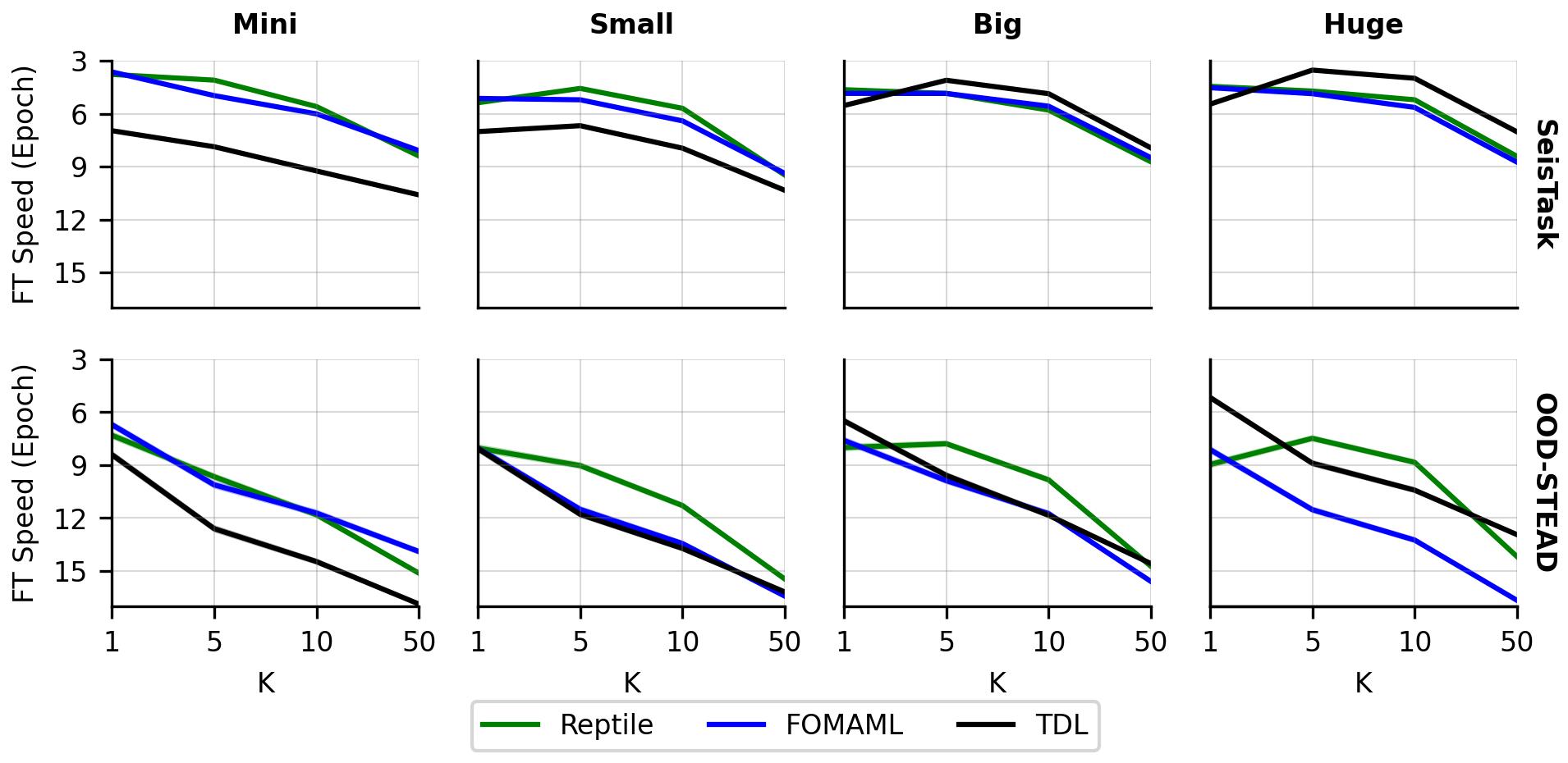}
        \caption{The fine-tuning speed (average number of fine-tuning epochs until the maximum performance on the evaluation data is reached) is shown as a function of the amount of fine-tuning data $K$. These results are aggregated over $N$ due to similarity in trends. Columns are the architecture size and rows are the evaluation task-sets. Meta-learning algorithms tend to reach their best performance faster for smaller architectures, and the results are mixed for larger architectures.}
        \label{fig:loss_speed}
    \end{figure}
    
    \begin{figure}[H]
        \centering
        \includegraphics[]{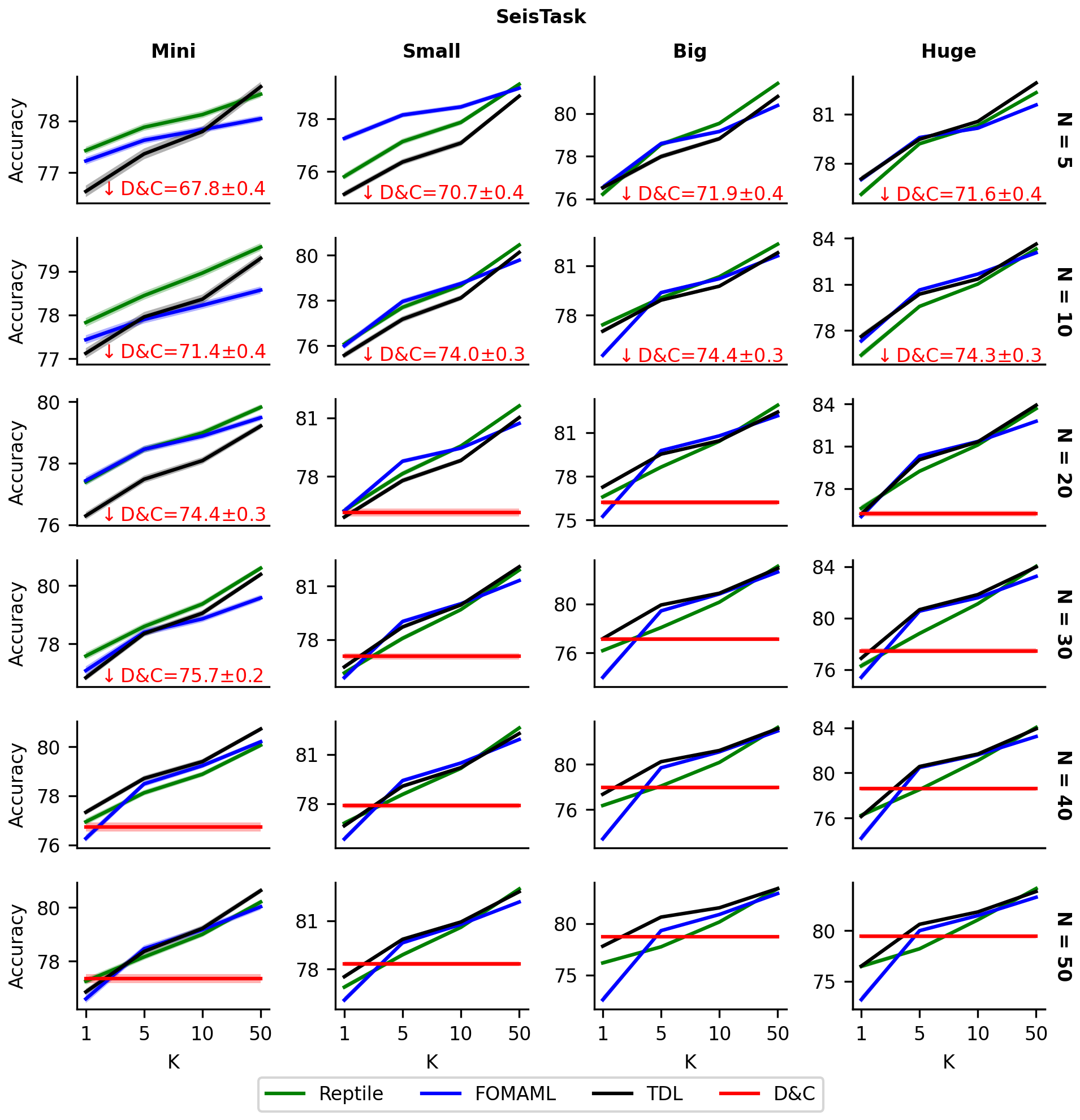}
        \caption{Average accuracy on \textbf{SeisTask} as a function of the amount of fine-tuning data $K$. Columns are architecture size and rows are the amount of training data $N$. Meta-learning typically exhibits as good or better performance than TDL except when the architecture size and the amount of training data grow. In addition, all algorithms tend to outperform the D\&C approach for non-trivial $K=1$, suggesting the knowledge learned from training/validation splits of SeisTask is highly transferrable to the SeisTask test split.}
        \label{fig:acc_seistask}
    \end{figure}

\end{document}